\documentclass[runningheads]{llncs}

\usepackage[mobile]{eccv}

\usepackage{eccvabbrv}

\usepackage{graphicx}
\usepackage{booktabs}
\usepackage{float} 
\usepackage{algorithm,algorithmic}

\usepackage[accsupp]{axessibility}  

\usepackage[pagebackref,breaklinks,colorlinks,citecolor=eccvblue]{hyperref}

\usepackage{orcidlink}

\newcommand{\bd}{\boldsymbol}
\newcommand{\E}{\mathbb{E}}
\newcommand{\dd}{\mathrm{d}}
\newcommand{\Normal}{\mathcal{N}(\bd{0},\bd{I})}
\newcommand{\Lsds}{L_{\text{SDS}}^\theta}
\newcommand{\Lfsd}{L_{\text{FSD}}^\theta}
\newcommand{\lsds}{L_{\text{sds}}^\theta}
\newcommand{\lsdsdd}{L_{\text{sds-2d}}^\theta}
\newcommand{\gc}{\bd{g}_\theta(\bd{c})}
\newcommand{\ec}{\bd{\epsilon}(\bd{c})}
\newcommand{\xo}{\hat{\bd{x}}^{\text{c}}_t}
\newcommand{\xc}{\hat{\bd{x}}^{\text{gt}}_t}
\newcommand{\epsf}{\tilde{\bd{\epsilon}}}
\newcommand{\window}{\bd{W}_{(W\frac{\phi_{\text{cam}}}{2\pi},H\frac{\theta_{\text{cam}}}{\pi})}}

\begin{document}

\title{Flow Score Distillation for Diverse Text-to-3D Generation}


\author{Runjie Yan\inst{1} \and
Kailu Wu\inst{2} \and
Kaisheng Ma\inst{2}\thanks{Corresponding author}}

\authorrunning{Runjie Yan et al.}

\institute{Institute for Interdisciplinary Information Sciences, Tsinghua University\and
ArchipLab, Institute for Interdisciplinary Information Sciences, Tsinghua University\\
\email{\{yanrj21, wkl22\}@mails.tsinghua.edu.cn}\\
\email{kaisheng@mail.tsinghua.edu.cn}}

\maketitle

\begin{abstract}
  Recent advancements in Text-to-3D generation have yielded remarkable progress, particularly through methods that rely on Score Distillation Sampling (SDS). While SDS exhibits the capability to create impressive 3D assets, it is hindered by its inherent maximum-likelihood-seeking essence, resulting in limited diversity in generation outcomes. In this paper, we discover that the Denoise Diffusion Implicit Models (DDIM) generation process (\ie PF-ODE) can be succinctly expressed using an analogue of SDS loss. One step further, one can see SDS as a generalized DDIM generation process. Following this insight, we show that the noise sampling strategy in the noise addition stage significantly restricts the diversity of generation results. To address this limitation, we present an innovative noise sampling approach and introduce a novel text-to-3D method called Flow Score Distillation (FSD). Our validation experiments across various text-to-image Diffusion Models demonstrate that FSD substantially enhances generation diversity without compromising quality.
\end{abstract}

\section{Introduction}\label{sec:intro}
In the realm of 3D content creation, a crucial step within the modern game and media industry involves crafting intricate 3D assets. Recently, 3D generation has facilitated the creation of 3D assets with ease. 3D generative models could be trained directly on certain representations (\eg point clouds\cite{achlioptas2018learning, luo2021diffusion}, voxel\cite{xie2018learning, smith2017improved} and mesh\cite{zhang2021sketch2model}). However, despite the recent efforts of Ojaverse\cite{deitke2023objaverse}, 3D data remains relatively scarce, especially when compared to the abundant 2D image data available on the internet. This scarcity constrains the generative capabilities of models trained solely on 3D datasets. Notably, the most prevailing text-to-3D approach is based on Score Distillation Sampling (SDS), proposed by Dreamfusion\cite{poole2022dreamfusion} and SJC\cite{wang2023score}. SDS effectively tackles the scarcity of 3D data by leveraging pretrained 2D text-to-image Diffusion Models, without directly training models on 3D datasets.

\begin{figure}[H]
  \centering
  \includegraphics[width=\linewidth]{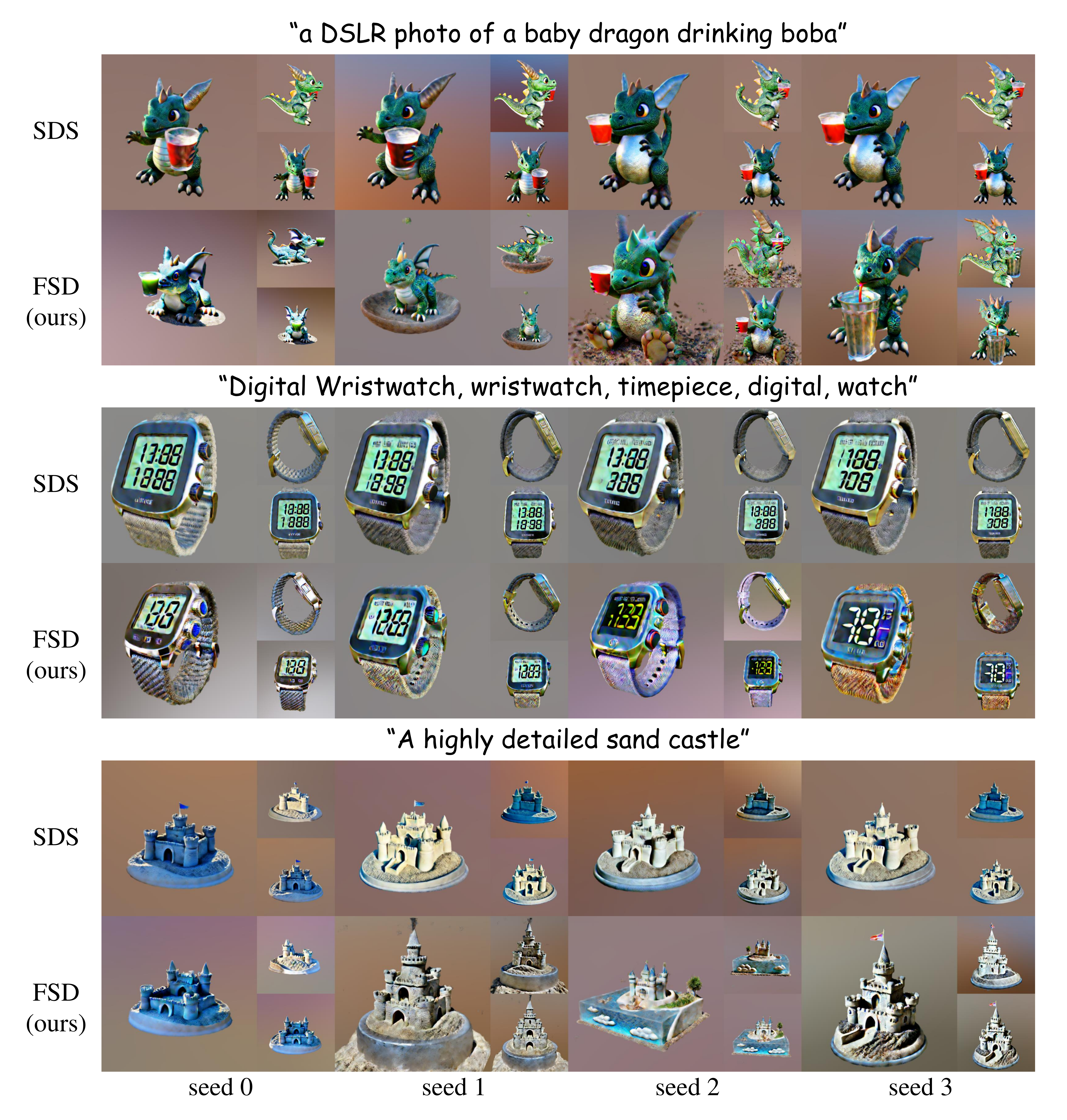}
  \caption{\textbf{Generation results of FSD and baseline method SDS.} \textbf{FSD} uses pretrained text-to-image Diffusion Models to generate realistic 3D models from text prompts. We improve the noise sampling strategy upon SDS and achieve \textbf{diverse generation results without compromising quality}.
  }
  \label{fig:intro}
\end{figure}

\begin{figure}[tb]
  \centering
  \includegraphics[width=\linewidth]{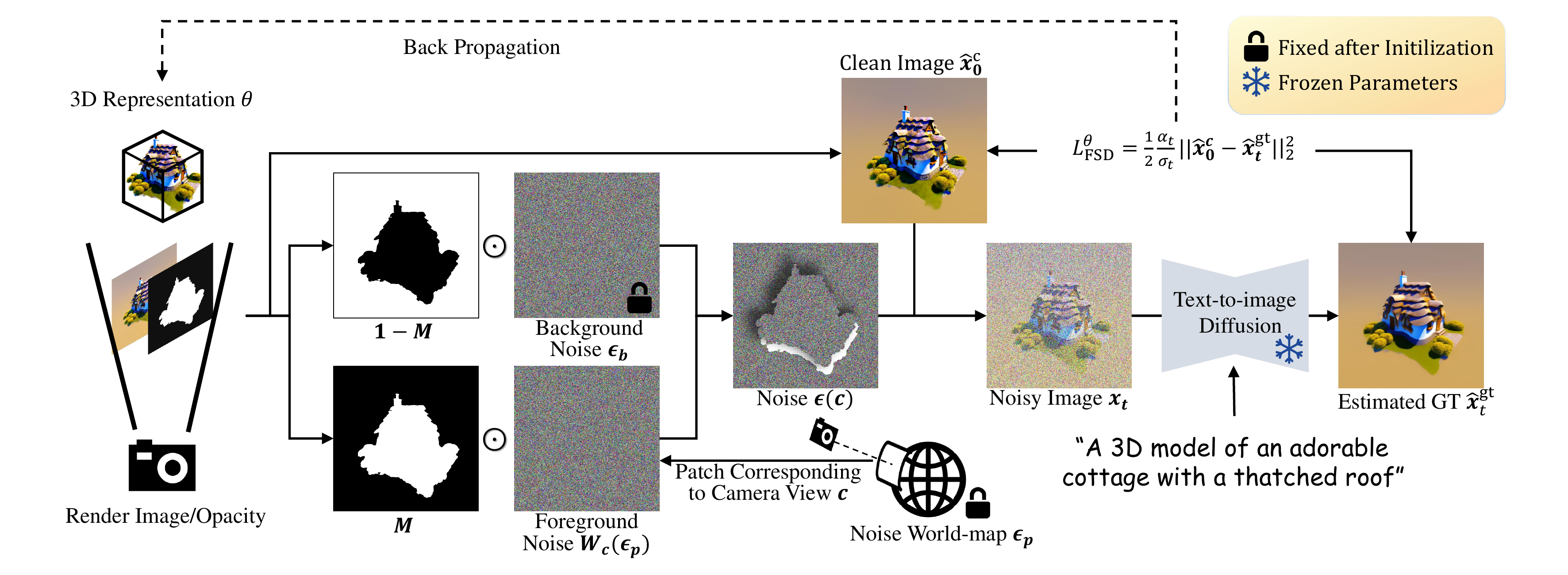}
  \caption{\textbf{Methods overview of FSD.} We propose Flow Score Distillation for text-to-3D generation by lifting a pretrained Diffusion Model. FSD renders an image $\gc$ from the 3D representation and adds noise  $\ec$ to the rendered image. To compute parameter updates according to $\Lfsd$, FSD uses a frozen text-to-image Diffusion Model to predict the noise $\ec$ added on image $\gc$. Similar to SDS\cite{poole2022dreamfusion, wang2023score}, FSD computes $\Lfsd$ by an image reconstruction loss between the ``clean image'' $\xo = \gc$ and ``ground-truth image'' $\hat{\bd{x}}_0$ predicted by the pretrained Diffusion Model. FSD further adopts timestep annealing schedule and noise sampling strategy. Instead of sampling noise from Gaussian distribution at each step of the optimization like SDS, we generate noise according to the deterministic noise function $\ec$, which is determined at the beginning of the optimization. 
  }
  \label{fig:methods}
\end{figure}

SDS is designed to optimize any representations (\eg Neural Radiance Field\cite{mildenhall2021nerf, muller2022instant,wang2021neus}, 3D Gaussian Splatting\cite{kerbl20233d}, Mesh\cite{Laine2020diffrast, shen2021dmtet} or even 2D images) that could render 2D images through probability density distillation\cite{oord2018parallel} using the learned score functions from the Diffusion Models. One of the main limitations of current SDS-based methods is that their distillation objectives will maximize the likelihood of the image rendered from the 3D representations, which leads to limited diversity. Additionally, The success of SDS relies heavily on large classifier-free guidance (CFG)\cite{ho2022classifier, yu2023text}. However, this reliance introduces issues like over-saturation and further restricts the diversity of the generation results. Despite several subsequent efforts\cite{wang2024prolificdreamer, zhu2023hifa, liang2023luciddreamer, katzir2023noise, huang2023dreamtime, tang2023stable, wang2023taming, armandpour2023re} to enhance SDS, the maximum-likelihood-seeking essence of the method remains unchanged. Notably, ProlificDreamer\cite{wang2024prolificdreamer} introduces Variational Score Distillation (VSD) and uses a fine-tuned Diffusion Model to model distribution on particles, which could alleviate the maximum-likelihood-seeking issues. However, training costs could grow linearly with the particle number of VSD. ESD\cite{wang2023taming} also points out that single-particle VSD is equivalent to SDS, which remains rooted in the essence of maximum likelihood seeking.

In this paper, we present a fresh perspective on SDS by viewing it as a generalized DDIM\cite{song2020denoising} generation process for 3D representations. Specifically, we discovered that the DDIM generation process (\ie PF-ODE \cite{song2020score}) can be succinctly expressed using an analogue of SDS loss. Surprisingly, by studying the difference between the analogue of SDS loss and the original form of SDS loss, we find that the noise sampling strategy during the noise addition stage appears to be the main cause that drives SDS toward mode-seeking behavior. SDS-based methods typically use random noise sampled from a Gaussian distribution at each optimization step, following the proposal of Dreamfusion\cite{poole2022dreamfusion} and SJC\cite{song2020score}. However, the variation in sampled noise can lead to varied optimization directions, which may harm the performance of SDS, as observed by ISM\cite{liang2023luciddreamer}. As we will show in this work, PF-ODE can be expressed by an analogue of SDS loss that uses the same noise throughout the generation process, which is different from the original proposal of SDS\cite{poole2022dreamfusion,wang2023score}. Based on this insight, we propose a novel noise sampling strategy to align SDS with the DDIM generation process on 3D representations.

This paper aims to overcome the aforementioned diversity challenge of SDS. We will first reveal an underlying connection between SDS and DDIM on 2D image generation. We will also show that the noise sampling strategy could be the primary factor that leads to the restricted diversity. Based on this insight, we will give our interpretation of SDS. From our novel viewpoint on SDS, we propose a novel approach called \textit{Flow Score Distillation} (FSD). FSD improves SDS by using a carefully designed noise sampling strategy. We lift our observations on image generation with FSD to 3D by proposing a view-dependent noise function $\ec$. We conduct validation experiments across various 2D Diffusion Models and demonstrate that FSD can achieve diverse generation outcomes without compromising quality as well as introducing extra training costs. Overall, our contributions can be summarized as follows.
\begin{itemize}
  \item We provide an in-depth analysis of SDS, an effective method in text-to-3D
  generation. We present a proposition that reveals the underlying connection between SDS and DDIM. Specifically, the DDIM generation process (\ie PF-ODE) can be succinctly expressed using an analogue of SDS loss. As a result, we can interpret SDS as a generalized DDIM generation process on 3D representations.
  \item Building upon our new insight into SDS, we identify that the diversity degradation stems from the noise sampling strategy. We show that adding the same noise throughout the generative process can enhance both generation quality and diversity in 2D generation with SDS. Hence, we show the noise sampling strategy is the primary factor that restricts the diversity.
  \item We introduce FSD as a cheap but effective solution to tackle the diversity challenges arising from the maximum-likelihood-seeking nature of SDS. We propose a deterministic \textit{world-map noise function} to generate coarsely aligned noise in 3D space. By applying a reasonable noise sampling strategy, FSD breaks free from the mode-seeking nature of SDS-like methods. Our validation experiments demonstrate that FSD substantially enhances generation diversity without compromising quality.
\end{itemize}

\section{Preliminaries and Related Works}

\label{sec:background}
In this section, we will provide preliminaries on Diffusion Models, differential equations associated with them, and SDS. Additionally, we will also introduce works related to this paper. Our notation conventions mainly follow Song \etal\cite{song2020score}, Lu \etal\cite{lu2022dpm} and Poole \etal\cite{poole2022dreamfusion}.

\subsection{Diffusion Models}

Diffusion Models\cite{sohl2015deep, ho2020denoising, song2020score} are a family of powerful generative models that are trained to gradually transform Gaussian noise to samples from a target distribution $p_0$. Their generation ability is further enhanced given datasets comprising billions of image-text pairs\cite{changpinyo2021conceptual, schuhmann2022laion, sharma2018conceptual}.

Assume the target distribution is $p_0$ and condition $y$. Diffusion Models define a forward process $\{x_t\}_{t\in[0,T]}$ starting from $\bd{x}_0 \sim p_0(\cdot|y)$, such that for any $t\in[0,T]$ the distribution of $x_t$ conditioned on $x_0$ satisfies:
\begin{equation}
  \bd{x}_t = \alpha_t \bd{x}_0 + \sigma_t \bd{\epsilon},
  \quad
  \bd{\epsilon} \sim \Normal
  ,\quad
  \bd{x}_0 \sim p_0(\cdot|y),\label{eq:forward-dpm}
\end{equation}
where $\alpha_t,\sigma_t\in\mathbb{R}^+$ are functions of $t$, defined by the \textit{noise schedule} of the Diffusion Model. And the (noisy) distribution at timestep $t$ is noted as $p_t$. 

In practice, a Diffusion Model is a neural network $\bd{\epsilon}_\phi(\bd{x}_t|y,t)$ parameterized by $\phi$ and is trained by minimizing the following objective\cite{song2020denoising, song2020score}:
\begin{equation}
  L_{\text{DMs}}^\phi = \frac{1}{2}\E_{
      \bd{x}_0\sim p_0(\cdot|y),
      \bd{\epsilon},
      t
    } \left[w_t ||\bd{\epsilon}_\phi(\bd{x}_t|y,t) - \bd{\epsilon}||^2_2\right],\label{eq:objective-dpm}
\end{equation}
where $w_t$ is a weighting function. As the training objective implies, a Diffusion Model can be seen as predicting the Gaussian noise added to the clean data $\bd{x}_0\sim p_0(\cdot|y)$. Song \etal\cite{song2020score} proved that:
\begin{equation}
  \bd{\epsilon}_\phi(\bd{x}_t|y,t) \approx - \sigma_t \nabla_{\bd{x}} \log p_t(\bd{x}_t|y), \label{eq:score}
\end{equation}
if the Diffusion Model $\bd{\epsilon}_\phi$ is trained to almost optimum. The term $\nabla_{\bd{x}} \log p_t(\bd{x}_t|y)$ in the above equation is also known as the \textit{score function}. 

\subsection{Diffusion PF-ODE and DDIM}
To generate samples from Diffusion Models, there exist different methods among the Diffusion Models family. Denoise Diffusion Implicit Models (DDIM)\cite{song2020denoising} designed a deterministic method for fast sampling from Diffusion Models. Recent works \cite{salimans2022progressive,karras2022elucidating,lu2022dpm} have shown that the sampling algorithm of DDIM is a first-order discretization of the Diffusion Probability Flow Ordinary Differential Equation (PF-ODE)\cite{song2020score}. Therefore, in this work, we will not make specific distinction between DDIM generation process and PF-ODE.

Theoretically, Diffusion PF-ODE yields the same marginal distribution as the forward process of Diffusion Models (\cref{eq:forward-dpm})\cite{song2020score}. We can write Diffusion PF-ODE\cite{karras2022elucidating, song2020score} (see detailed derivation in Appendix) as:
\begin{align}
  \frac{\dd (\bd{x}_t/\alpha_t) }{\dd t} 
  &= \frac{\dd (\sigma_t/\alpha_t)}{\dd t} \left( -\sigma_t \nabla_{\bd{x}} \log p_t(\bd{x}_t|y) \right)\\
  &= \frac{\dd (\sigma_t/\alpha_t)}{\dd t} \bd{\epsilon}_\phi(\bd{x}_t|y,t),
  \quad
  \bd{x}_T \sim p_T(\bd{x}_T|y). \label{eq:pf-ode-simple}
\end{align}
Notably, one can generate a sample $\bd{x}_0$ in the target distribution $p_0(\bd{x}_0|y)$ by following the PF-ODE trajectory from $t=T$ to $t=0$, starting from $\bd{x}_T \sim p_T(\bd{x}_T|y)=\Normal$. 

\subsection{Score Distillation Sampling}
Recently, DreamFusion\cite{poole2022dreamfusion} and SJC\cite{wang2023score} proposed Score Distillation Sampling (SDS) to generate 3D models by optimizing a differentiable 3D representation using priors from text-to-image Diffusion Models. Follow-up works tried to improve upon SDS through various aspects, \eg coarse-to-fine training strategy\cite{lin2023magic3d, wang2024prolificdreamer, chen2023fantasia3d}, disentangled 2D-3D priors\cite{chen2023fantasia3d, ma2023geodream, wang2024prolificdreamer} and refined formulas\cite{zhu2023hifa, wang2024prolificdreamer, liang2023luciddreamer, tang2023stable, wang2023taming, yu2023text, armandpour2023re}. Moreover, due to the lack of comprehensive 3D-aware knowledge, multi-face Janus problem often arises when using SDS \cite{poole2022dreamfusion}. To mitigate this challenge, one can consider replacing the text-to-image Diffusion Models with Diffusion Models designed for object novel view synthesis\cite{liu2023zero, long2023wonder3d, liu2023syncdreamer, weng2023consistent123, ye2023consistent} or multi-view Diffusion Models\cite{shi2023mvdream}. Such an adaptation can significantly alleviate the multi-face Janus problem encountered in 3D generation using SDS.

SDS is first introduced by DreamFusion\cite{poole2022dreamfusion} and SJC\cite{wang2023score} to apply image diffusion priors for 3D generation. SDS can optimize on any representations parameterized by $\theta$, which can render an image $\gc$, given camera parameter $\bd{c}$. Basically, SDS defines a probability density distillation\cite{oord2018parallel} loss, denoted as $\Lsds$, whose gradient writes as follows:
\begin{align}
  \nabla_\theta \Lsds
  =& \E_{\bd{c},t} \left[ w_t \frac{\sigma_t}{\alpha_t} \nabla_\theta D_{\text{KL}}\left(p_t(\bd{x}_t|\bd{x}_0=\gc) || p_t(\bd{x}_t|y)\right) \right] \label{eq:sds-kl}\\
  =& \E_{\bd{\epsilon},\bd{c},t}  \left[ w_t \left(\bd{\epsilon}_\phi(x_t|y,t)-\bd{\epsilon}\right)\frac{\partial \gc}{\partial \theta}\right], \label{eq:sds-sjc}
\end{align}
where ${\epsilon}_\phi$ is a text-to-image Diffusion Model and $y$ is the generation condition, \eg text prompts. In order to generate 3D content, SDS needs to go through an optimization process on the 3D representation parameter $\theta$ to generate a single 3D model. 

Even though SDS can produce high-fidelity objects, there has been ongoing debate about the underlying theory. Recent works \cite{shi2023mvdream,liang2023luciddreamer} also show that $\Lsds$ is equivalent to a reconstruction loss:
\begin{equation}
  \Lsds = \E_{\bd{\epsilon},\bd{c},t} \left[ \frac{1}{2} w_t \frac{\alpha_t}{\sigma_t}||\xo - \xc||_2^2\right] \label{eq:sds-x0},
\end{equation}
where $\xo=\gc$ and $\xc=\frac{\bd{x}_t-\sigma_t \bd{\epsilon}_\phi (\bd{x}_t|y,t)}{\alpha_t}$ is the one-step ``estimated ground-truth image'' from Diffusion Models, whose gradient is detached. Some other works\cite{yu2023text, katzir2023noise, tang2023stable} also tried to explain SDS by analyzing the function of each component of SDS loss. In this paper, we will provide another interpretation of SDS: it can be viewed as a generalized DDIM generation process.

Another simple yet effective technique is to apply timestep annealing trick\cite{huang2023dreamtime,wang2024prolificdreamer,zhu2023hifa}, which can improve generation quality significantly. This technique is intuitive because reducing added noise during the latter stages of optimization, enabling models to discern finer details and iteratively improve upon them. Let us denote the time in the SDS optimization process as $\tau$ and the term that was taken expectation in the definition of SDS (\cref{eq:sds-x0}) as $ \lsds (\bd{\epsilon}, \bd{c}, t) = \frac{1}{2} \frac{\alpha_t}{\sigma_t} ||\xo - \xc||_2^2 $. We use lowercase letters footnote for $\lsds$ to distinguish it from \cref{eq:sds-x0}. Finally, the SDS optimization process with timestep annealing can be written as:
\begin{equation}
  \frac{\dd \theta}{\dd \tau} = w_t \E_{\bd{\epsilon},\bd{c}} \left[ \nabla_\theta \lsds (\bd{\epsilon}, \bd{c}, t=t(\tau)) \right], \label{eq:sds-anneal}
\end{equation}
where $t(\tau)$ is a monotonically decreasing function of $\tau$.

\section{Flow Score Distillation for 2D Generation}
\label{sec:2D}

\begin{figure}[t]
  \centering
  \begin{subfigure}{0.18\linewidth}
    \includegraphics[width=\linewidth]{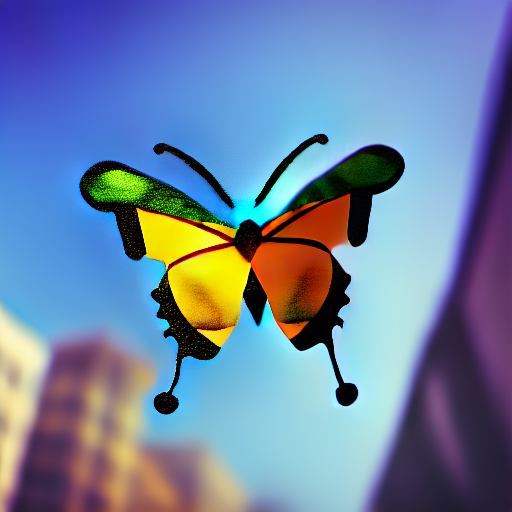}\\
    \includegraphics[width=\linewidth]{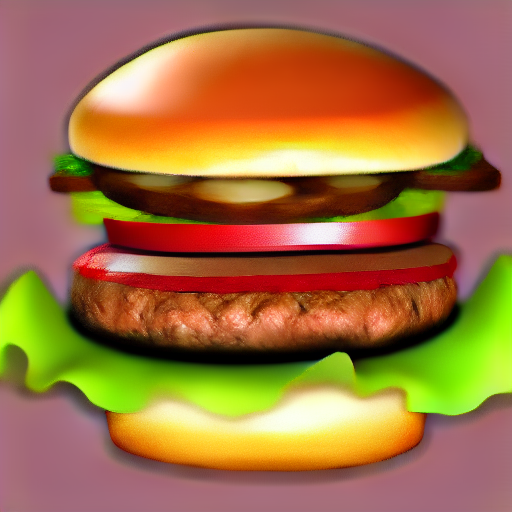}\\
    \includegraphics[width=\linewidth]{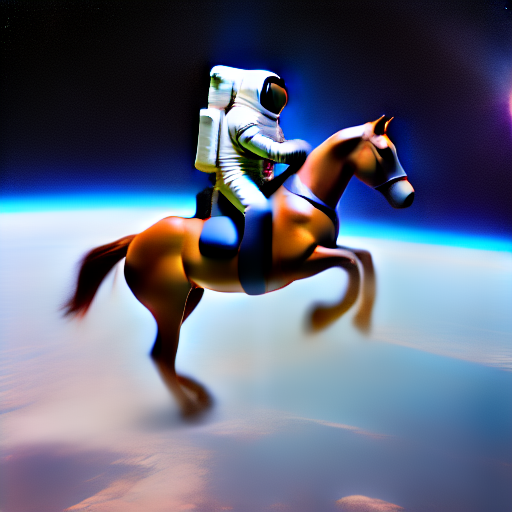}
    \caption*{SDS\cite{poole2022dreamfusion, wang2023score}}
  \end{subfigure}
  \hfill
  \begin{subfigure}{0.18\linewidth}
    \includegraphics[width=\linewidth]{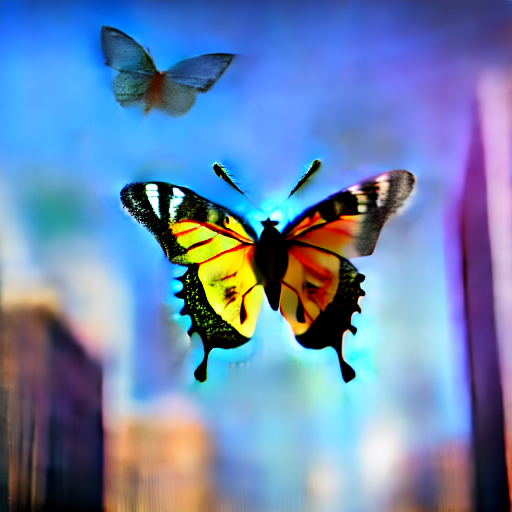}\\
    \includegraphics[width=\linewidth]{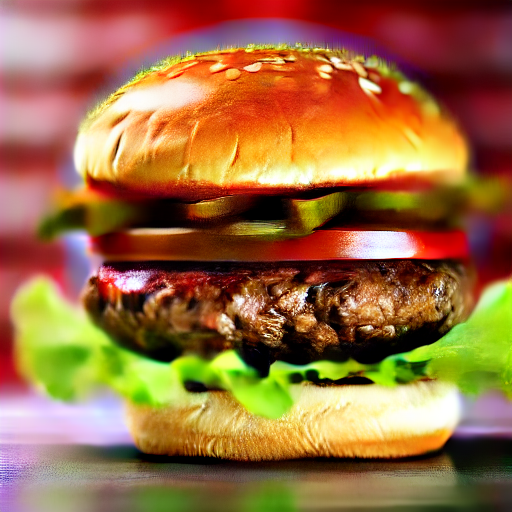}\\
    \includegraphics[width=\linewidth]{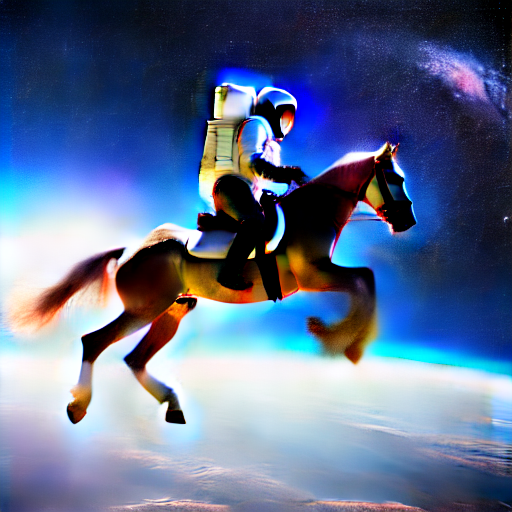}
    \caption*{NFSD\cite{katzir2023noise}}
  \end{subfigure}
  \hfill
  \begin{subfigure}{0.18\linewidth}
    \includegraphics[width=\linewidth]{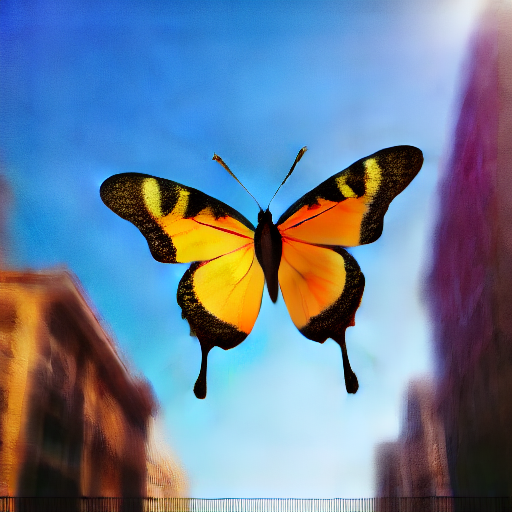}\\
    \includegraphics[width=\linewidth]{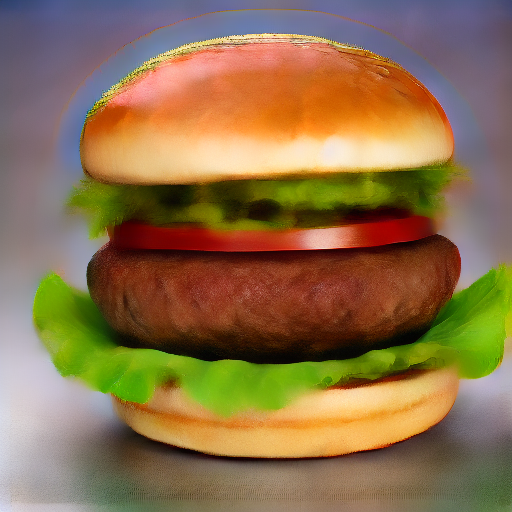}\\
    \includegraphics[width=\linewidth]{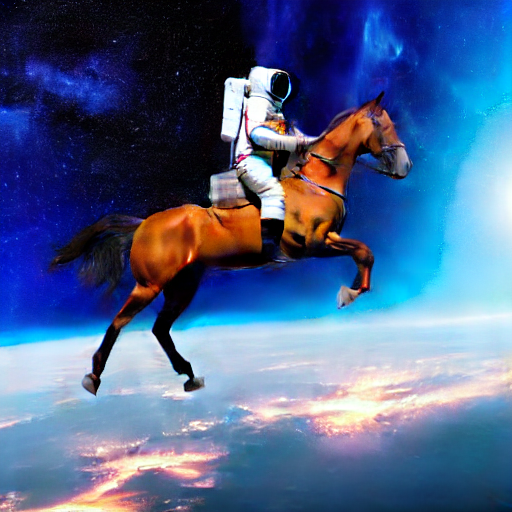}
    \caption*{VSD\cite{wang2024prolificdreamer}}
  \end{subfigure}
  \hfill
  \begin{subfigure}{0.18\linewidth}
    \includegraphics[width=\linewidth]{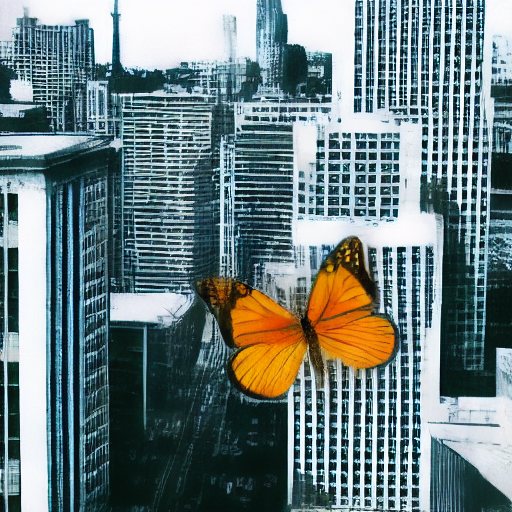}\\
    \includegraphics[width=\linewidth]{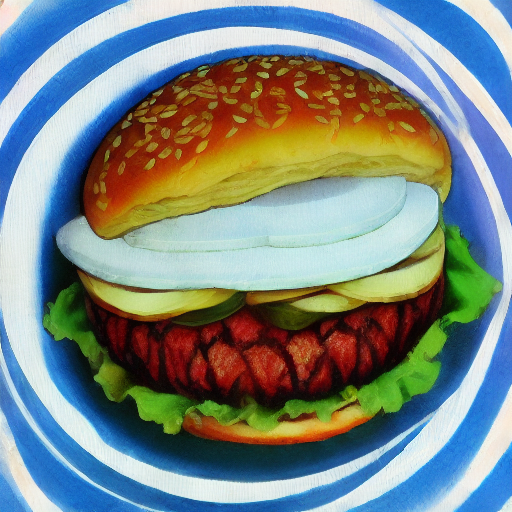}\\
    \includegraphics[width=\linewidth]{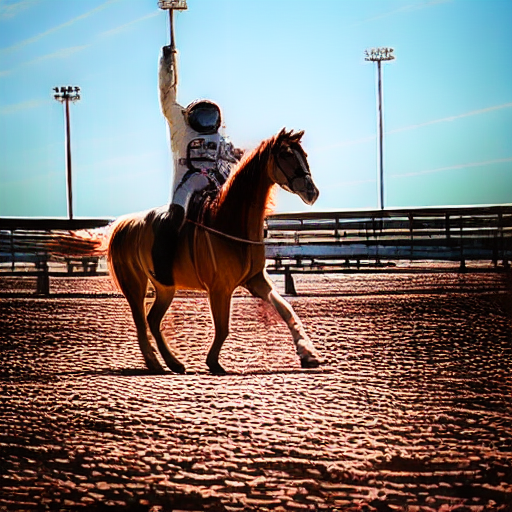}
    \caption*{FSD (ours)}
  \end{subfigure}
  \quad\;
  \begin{subfigure}{0.18\linewidth}
    \includegraphics[width=\linewidth]{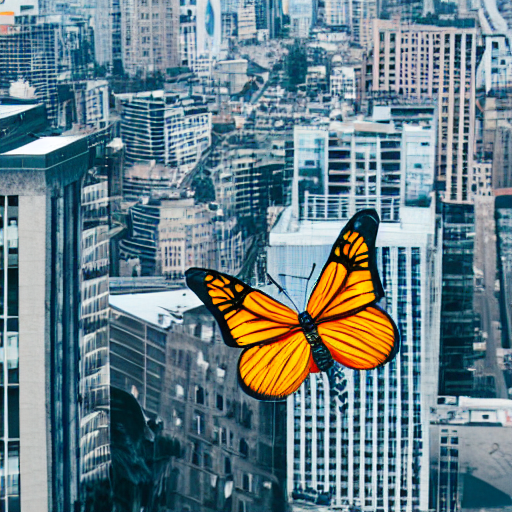}\\
    \includegraphics[width=\linewidth]{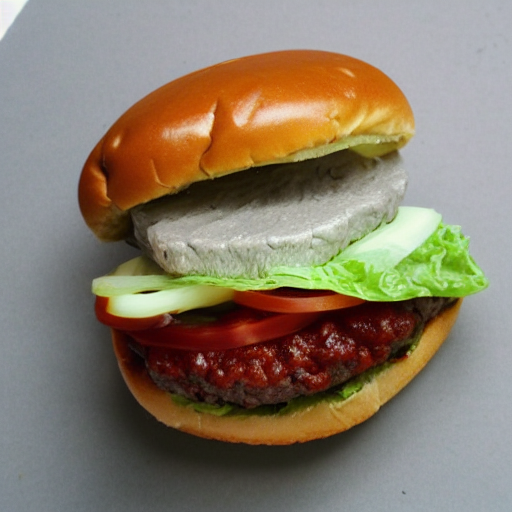}\\
    \includegraphics[width=\linewidth]{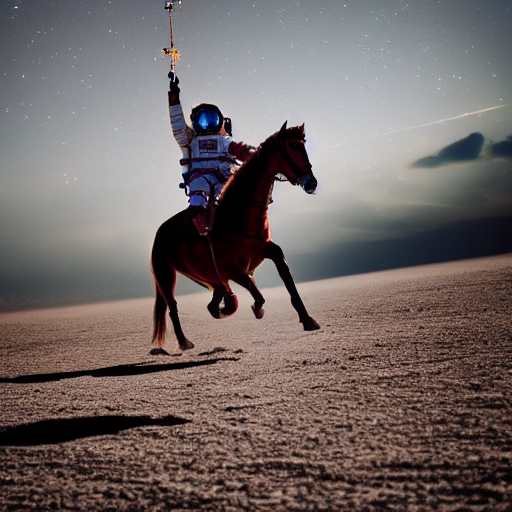}
    \caption*{DDIM\cite{song2020denoising}}
  \end{subfigure}
  \caption{\textbf{Generation results of different methods on image space with the same random seeds.} FSD can generate images that 
 are very similar to images generated by DDIM given the same initial noise (implied by \cref{proposition:fsd}). However, FSD can also be used for 3D generation, a task for which DDIM is not suitable. See experiment details in Appendix.
  }
  \label{fig:fsd-2d}
\end{figure}

In this section, we only consider generation on 2D using SDS as SDS loss can also be applied to image representations. In this case, $\theta=\gc$ and $\frac{\partial \gc}{\partial \theta} = \bd{I}$. Then $\lsds$ becomes the following form:
\begin{equation}
  \nabla_\theta \lsdsdd (\bd{\epsilon}, t) = \bd{\epsilon}_\phi(x_t|y,t)-\bd{\epsilon} \label{eq:tiny-sds-2d}.
\end{equation}
We will reveal a simple but profound connection between SDS and DDIM and give our interpretation of SDS in this section.

\subsection{Simplified Formulation of Diffusion PF-ODE}
We first reveal that PF-ODE (\cref{eq:pf-ode-simple}) can be formulated by an analogue of SDS (\cref{eq:tiny-sds-2d}) in this section. We first define 
\begin{equation}
  \bd{x}_t = \alpha_t \xo + \sigma_t \epsf \label{eq:change-of-variable},
\end{equation}
where $\epsf$ is a constant for each ODE trajectory. Notice that when $t=T$, the initial condition of the ODE gives $ \epsf = 0 \cdot \xo + 1 \cdot \epsf = \bd{x}_T \sim \Normal$. Intuitively, \textbf{$\epsf$ can be viewed as the noise added to $\xo$, and $\xo$ as the clean image at timestep $t$}. So we will also refer to $\epsf$ as \textit{the initial noise} apart from \textit{the added noise} in this paper. It is noteworthy that the concept of the clean image $\xo=\frac{\bd{x}_t-\sigma_t \epsf}{\alpha_t}$ is different from the aforementioned estimated ground-truth image $\xc=\frac{\bd{x}_t-\sigma_t \bd{\epsilon}_\phi(\bd{x}_t|y,t)}{\alpha_t}$.  By applying ``\textit{change-of-variable}'' trick and change the variable of the Diffusion PF-ODE from $\bd{x}_t$ to $\xo$, we have:
\begin{proposition}[An equivalent form of Diffusion PF-ODE]
  \label{proposition:fsd}
  Diffusion PF-ODE (\cref{eq:pf-ode-simple}) can be equivalently formulated by an analogue of SDS loss (\cref{eq:tiny-sds-2d}):
  \begin{align}
    \frac{\dd \xo }{\dd t} 
    &= \frac{\dd (\sigma_t/\alpha_t)}{\dd t} \left[ \bd{\epsilon}_\phi(\bd{x}_t|t, y) - \epsf \right] \label{eq:pf-ode-sds}\\
    &= w_t' \nabla_\theta \lsdsdd (\epsf, t),
  \end{align}
where $\bd{x}_t = \alpha_t \xo + \sigma_t \epsf$, $\theta = \xo$ and $w_t' = \frac{\dd (\sigma_t/\alpha_t)}{\dd t} $ is a weighting scalar.
\end{proposition}
Please refer to Appendix for detailed derivation of this proposition. Remarkably, in the context of image generation, we observe that the evolution direction of PF-ODE aligns precisely with the gradient of the SDS loss (\cref{eq:tiny-sds-2d}).

\subsection{Flow Score Distillation on 2D}
\label{subsec:fsd-2d}
Even though we found the evolution direction of Diffusion PF-ODE (\cref{eq:pf-ode-sds}) is very samilar to the SDS loss, there exist some notable differences compared to the original definition of SDS loss (\cref{eq:sds-kl}). 
Specifically, 
i) the timestep in a DDIM process is monotonically decreasing, aligning with the timestep annealing technique\cite{wang2024prolificdreamer,zhu2023hifa,huang2023dreamtime}. But SDS uses randomly sampled timestep.
ii) And the change-of-variable trick (\cref{eq:change-of-variable}) we used during our simplification process implies we should also add the same noise $\epsf$ throughout the SDS generation process, to align it with DDIM. In contrast, the original SDS uses uncorrelated random noise.

As we will demonstrate in subsequent sections and through our experiments, the second difference between DDIM and SDS significantly influences generation diversity. Therefore, we term our approach that combines \textbf{timestep annealing and consistent noise sampling strategy throughout the generation process} as \textit{Flow Score Distillation} (FSD) to differ it from SDS. We visualize image generation results using several SDS-like methods, FSD and DDIM in \cref{fig:fsd-2d} to demonstrate the differences between SDS-based methods and FSD.

\subsection{Analysis of the Noise Sampling Strategy}
\label{subsec:init-noise}
\begin{figure}[t]
\centering
\includegraphics[width=\linewidth]{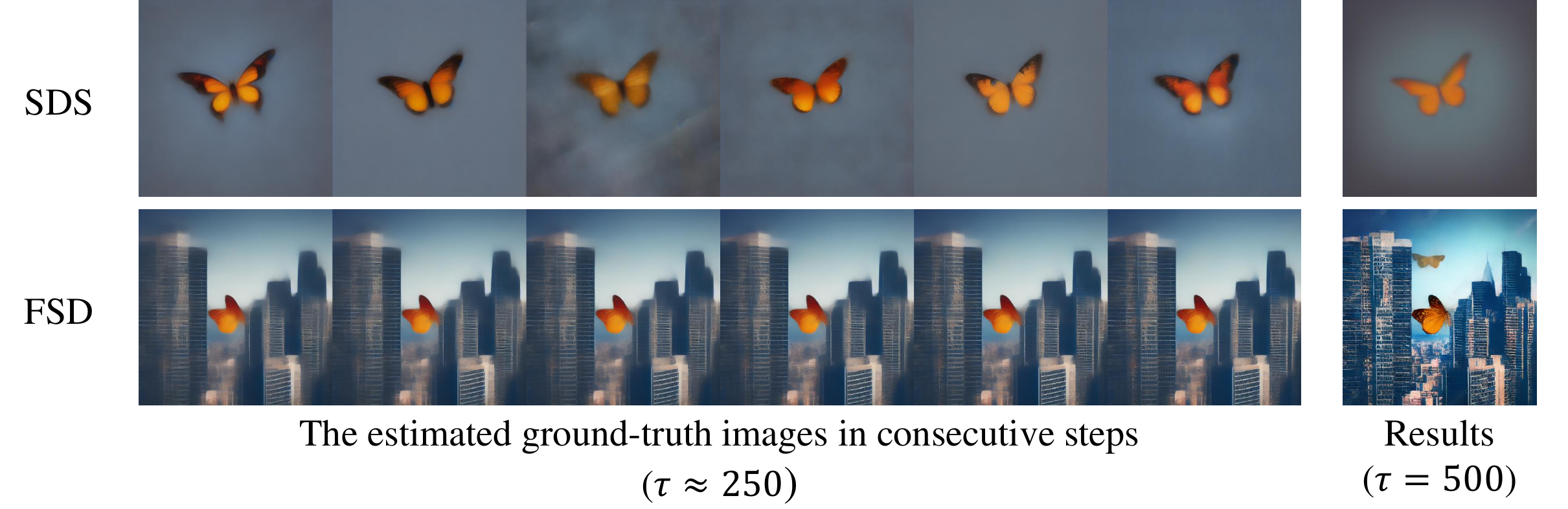}
  \caption{\textbf{Visualization of FSD and SDS for image generation.} We visualize generation results and the estimated ground-truth images in consecutive steps at halfway of the generation for both FSD and SDS\cite{poole2022dreamfusion, song2020score}. We find the estimated ground-truth images of FSD are consistent, while the contents vary greatly in the ground-truth images of SDS. We set CFG=7.5 and adopted the same linear timestep annealing schedule for both FSD and SDS in the experiment for this figure.
  }
  \label{fig:fsd-2d-analysis}
\end{figure}

Our empirical investigation reveals that the generation results produced by SDS exhibit an undesirable tendency toward over-smoothness and lack of diversity. Remarkably, even with the timestep annealing technique, this issue persists. This tendency originates from the maximum-likelihood-seeking nature implied by its definition (\cref{eq:sds-kl}) where it models the current distribution using a Dirichlet function centered at $\gc$. In contrast, not only does FSD yield diverse outcomes, but it can also generate highly detailed samples. This can be attributed to that FSD is more aligned with a DDIM process, which can generate samples from exactly the target distribution $p_0$. So we conclude that the noise sampling strategy can affect the generation diversity greatly.

To gain an intuitive understanding of these effects, we also explain the reasons by following the reconstruction loss interpretation \cite{shi2023mvdream,liang2023luciddreamer}. SDS loss can be seen as the reconstruction error between the rendered images and the one-step estimated ground-truth images predicted by the Diffusion Models. When different noise is added, the generated ground-truth images differ greatly (Demonstrated in \cref{fig:fsd-2d-analysis}), as noted by ISM\cite{liang2023luciddreamer}. As a result, SDS can only generate averaged outcomes. In contrast, applying identical noise at each optimization step ensures a more consistent optimization trajectory. So FSD can generate diverse results.

\subsection{Compare FSD with SDS and DDIM on 2D}
We summarize the difference between FSD, SDS, and DDIM when applied to 2D images as follows:

\subsubsection{Noise Sampling Strategy} 
Both FSD and DDIM add fixed noise sampled from $\Normal$ at the beginning of optimization to the clean image $\xo$. While SDS adds random noise sampled from $\Normal$ at each step of the optimization to the clean image $\xo$.

\subsubsection{Optimizer}
Both FSD and SDS use Adam\cite{kingma2014adam} (or Adan\cite{xie2022adan}) to update the clean image parameter $\xo$. Moreover, the impact of the weighting parameter $w_t$ is negligible as long as it changes slowly over $t$ since Adam optimizer is invariant to the scale of the gradients\cite{kingma2014adam}. While DDIM updates on $\bd{x}_t$ directly according to the first-order discretization of PF-ODE.

\subsubsection{Diffusion Timestep Schedule}
FSD tries to model the schedule of DDIM by using a monotonically decreasing function $t(\tau)$. While SDS uses time schedule which samples $t$ uniformly. Moreover, the convergence time $\tau_{\text{end}}$ of FSD and SDS is hard to determine theoretically, but DDIM will converge when $t=T$.

\section{Lifting Flow Score Distillation to 3D}
\label{sec:3D}

\begin{figure}[t]
\centering
\includegraphics[width=0.95\linewidth]{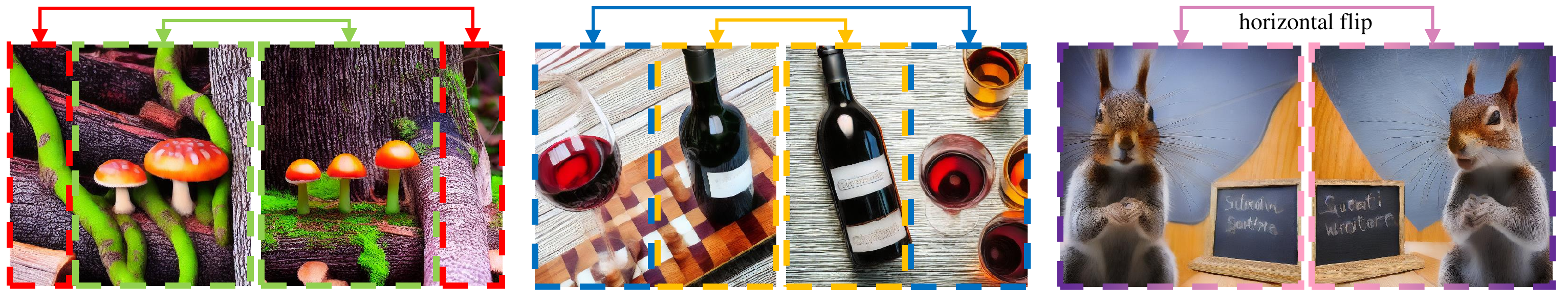}
  \caption{\textbf{Impact of initial noise $\epsf$.} Experiments show that the local textures of noise added during FSD optimization are highly correlated with the textures of the final image. We shuffle the patches of initial noise $\epsf$ used by FSD and observe that the textures of generated images are shuffled in the same way. This property inspired our design of world-map noise function $\ec$ for 3D generation in this work. In this figure, the parts framed by dotted lines of the same color share the same initial noise $\epsf$ patches.
  }
  \label{fig:fsd-2d-local}
\end{figure}
As highlighted in \cref{subsec:init-noise}, we have identified that the noise sampling strategy might contribute to the decline of the diversity of SDS significantly. Building upon this insight, we follow the discussion of FSD in \cref{subsec:fsd-2d} and propose to use deterministic noise generation strategy. We can directly generalize FSD to arbitrary 3D representations $\gc$:
\begin{align}
  \nabla_\theta\Lfsd 
  =& \E_{\bd{c}} \left[ \nabla_\theta\lsds (\bd{\epsilon}=\ec, \bd{c}, t=t(\tau)) \right]\\
  =& \E_{\bd{c}} \left[\left(\bd{\epsilon}_\phi(x_t|y,t(\tau))-\ec\right)\frac{\partial \gc}{\partial \theta} \label{eq:fsd} \right],
\end{align}
where $\bd{x}_t = \alpha_t \gc + \sigma_t \ec $, $\ec$ is a deterministic noise function generated at the beginning of the optimization and $t(\tau)$ is a monotonically decreasing timestep schedule function to optimization time $\tau$. We do not specify the form of the deterministic noise function $\ec$ in FSD. However, we propose some rules for designing $\ec$ based on the actual generation effect in Appendix, according to our practical experiences. We will also introduce the \textit{world-map noise function} as $\ec$ for our experiments of 3D generation with FSD in this paper.

Alternatively, FSD loss can be seen as applying DDIM generation process on 3D representations through Jacobian of a differentiable renderer. This viewpoint shares some similarities to the interpretation of SDS from SJC\cite{wang2023score}, who consider SDS as back-propagating the score\cite{song2020score} of Diffusion Models through Jocabian of the renderer. Meanwhile, $\ec$ is a deterministic noise function that generates correlated noise between views, which aligns with the prior that the nearby views are correlated. The design of FSD ensures the optimization directions are consistent, particularly when similar $\bd{c}$s are sampled. As the $\ec$s are similar, the generated ground-truth images should be consistent as well. Notably, recent works\cite{ge2023preserve, qiu2023freenoise, chang2024how} on video generation also find using designed video noise prior can improve the capabilities of Video Diffusion Models.

\subsection{Designing $\ec$.}
\label{subsec:noise-design}

\subsubsection{Failure of a Vanilla Design of $\ec$}
\label{subsubsec:noise-design-vanilla}

\begin{figure}[t]
  \centering
\includegraphics[width=0.8\linewidth]{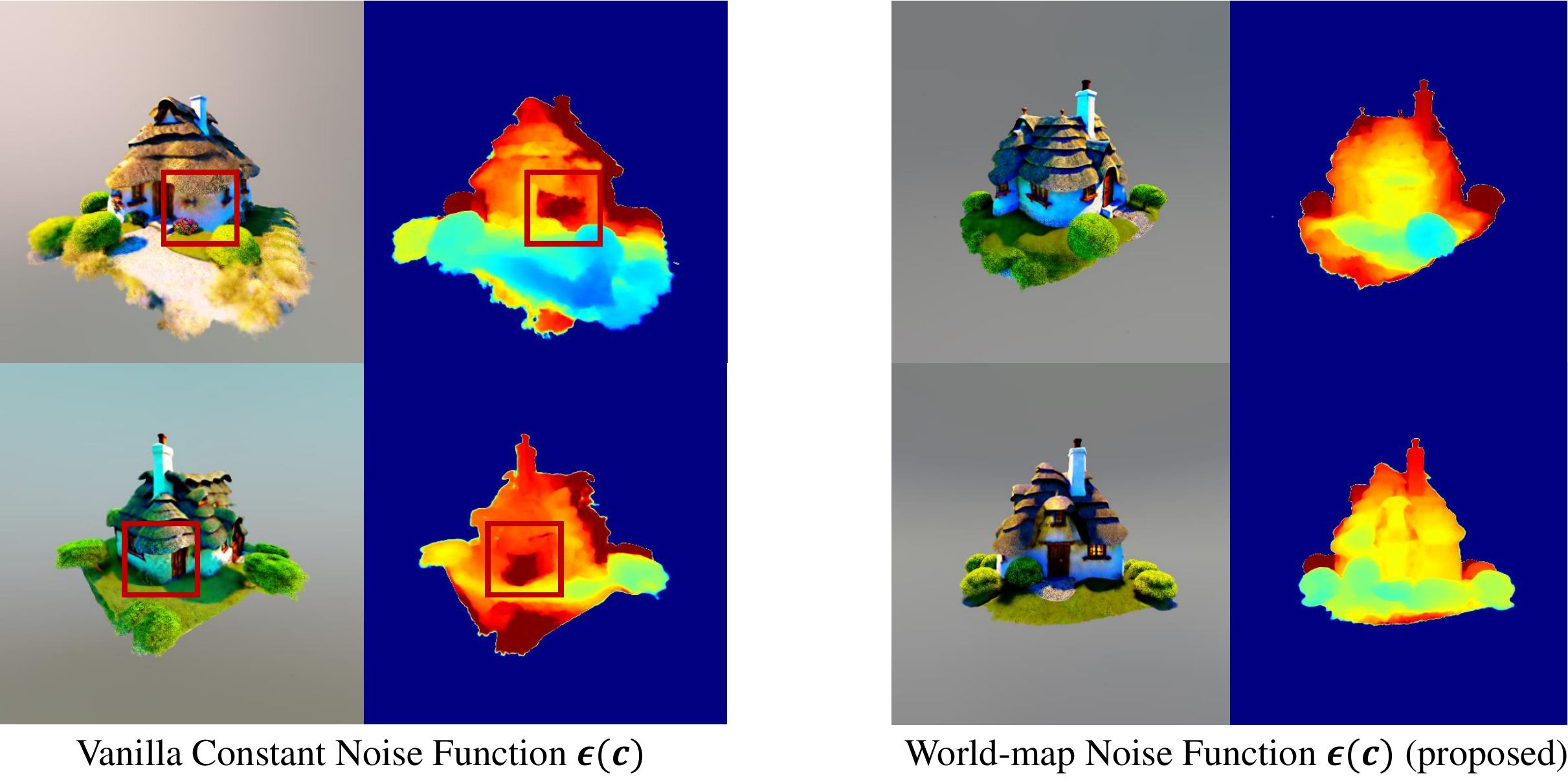}
  \caption{\textbf{Compare world-map noise function $\ec$ with a vanilla design of constant function $\bd{\epsilon}$.} We visualize the rendered images and depth maps for the two noise methods. The vanilla design of noise function $\bd{\epsilon}$ can easily lead to holes on the surfaces (framed in red), even when the RGB images seem plausible. In contrast, no obvious flaws are observed in the results of FSD with the world-map noise function.
  }
  \label{fig:failure}
\end{figure}

A vanilla design of $\ec$ can be $\ec = \bd{\epsilon}$, which is a constant function. However, according to our experiments on text-to-3D generation, such a design can lead to poor geometry of the generated samples. Typically, holes on the surfaces are observed (\cref{fig:failure}). We attribute this effect to the uneven convergence speed of FSD in 3D space caused by the constant noise function. Our experiments on 2D show that the local textures of noise added during FSD optimization are highly correlated with the local textures of the final image (Demonstrated in \cref{fig:fsd-2d-local}). And in text-to-3D generation with FSD, the generated ground-truth images have more consistent textures at the center point than other points in 3D space, due to the sampling strategy of camera view $\bd{c}$. As a result, the convergence speed at the center point is much higher than at other points, leading to holes on the surfaces. Flaws are also observed in Video Diffusion Models that adopt fixed noise prior, due to similar reasons, which is known as the textures sticking problem\cite{chang2024how}.

\subsubsection{World-map Noise Function  $\ec$.}
\label{subsubsec:noise-design-world-map}
To avoid relating specific noise textures to specific points in 3D space throughout the generation process but still augment the consistency of added noise between camera views, we propose \textit{world-map noise function} $\ec$ in this paper (visualized in \cref{fig:methods}), which aligns noise textures coarsely in 3D space while avoiding converging too fast at a specific point in 3D space (See discussion in Appendix).

Specifically, we first compute a foreground mask $\bd{M}(\bd{r}) = [\alpha(\bd{r}) > \alpha_0]$ for each ray $\bd{r}$, where $0\leq\alpha(r)$, $\alpha_0\leq1$ is the opacity of pixel $\bd{r}$. We add the same noise for the background and query a noise patch from the noise world-map $\bd{\epsilon}_p$ of size $D \times H \times W$ for the foreground. Let us denote the camera parameters sampled on the sphere as $\bd{c} = (\text{FOV}, r_{\text{cam}}, \theta_{\text{cam}},\phi_{\text{cam}})$. Both $\bd{\epsilon}_b$ and $\bd{\epsilon}_p$ are sampled from $\Normal$ at the beginning of the optimization. And $\bd{\epsilon}$ is re-sampled from $\Normal$ at each step of the optimization. The world-map noise function $\ec$ is:
\begin{equation}
  \ec = \sqrt{\beta} \cdot \left( (1-\bd{M}) \odot \bd{\epsilon}_b + \bd{M} \odot \window(\bd{\epsilon}_p) \right) + \sqrt{1-\beta} \cdot \bd{\epsilon} \label{eq:fsd-noise},
\end{equation}
where $\beta$ is a blending factor to balance between random noise and deterministic noise. If not specified, we set $\beta=1$ in this work so that $\ec$ is a completely deterministic noise function. $\window(\bd{\epsilon}_p)$ operation refers to the noise patch of size $D\times H_{\text{hidden}} \times W_{\text{hidden}}$ centered at position $(W\frac{\theta_{\text{cam}}}{2\pi},H\frac{\phi_{\text{cam}}}{\pi})$ on the noise worldmap $\bd{\epsilon}_p$. We also provide a pseudocode of our algorithm in Appendix.

\subsection{Compare FSD with SDS on 3D}
Apart from timestep annealing trick\cite{huang2023dreamtime, zhu2023hifa, wang2024prolificdreamer}, FSD is different from SDS in terms of noise sampling strategy as well. In case when the same camera view $\bd{c}$s are sampled, FSD yields consistent one-step estimated ground-truth images since $\ec$ is the same. Even when different camera views $\bd{c}$s are sampled, the ground-truth images are still coarsely aligned. In contrast, one can see SDS as using an uncorrelated noise prior $\ec$ on $\bd{c}$, which always yields ground-truth images that are inconsistent and have notable differences. We also discuss the differences between our method with recent works~\cite{wu2024consistent3d, gu2023boot} in \cref{app:sec:discussion}.
\section{Experiments}
\begin{figure}[t]
  \centering
  \begin{subfigure}{\linewidth}
  \centering
    \includegraphics[width=\linewidth]{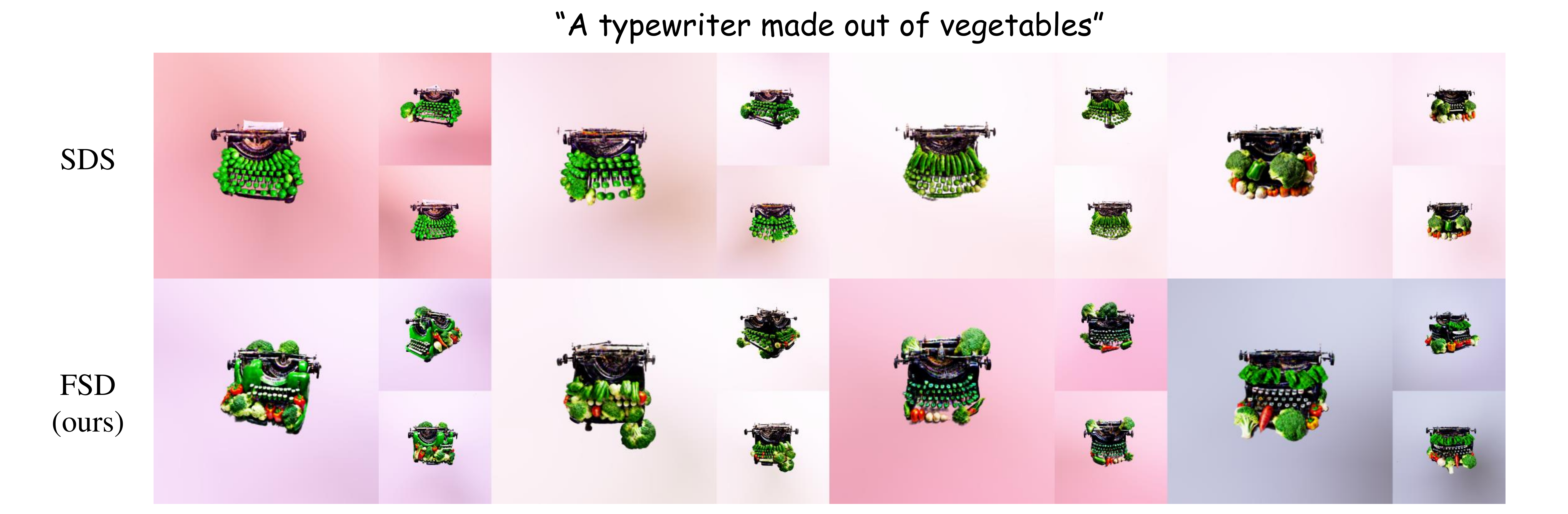}
    \caption{Stable Diffusion\cite{song2020denoising} as backbone Diffusion Model}\label{subfig:3d-sd}
  \end{subfigure}\\
  \begin{subfigure}{\linewidth}
  \centering
    \includegraphics[width=\linewidth]{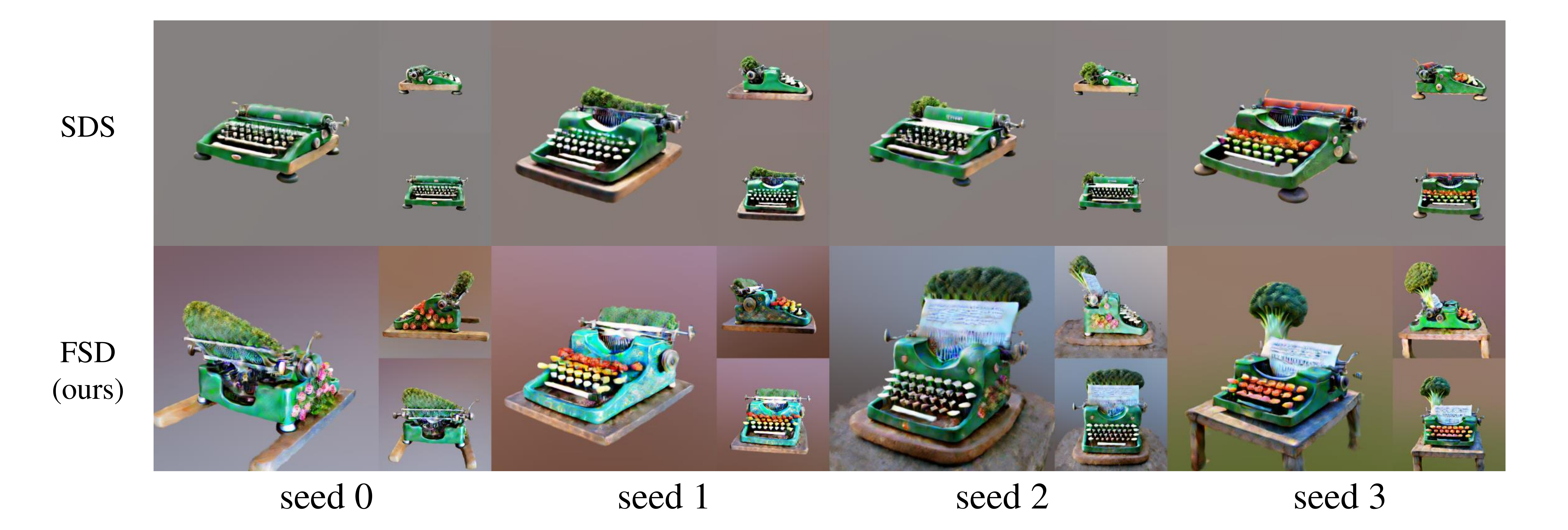}
    \caption{MVDream\cite{shi2023mvdream} as backbone Diffusion Model}\label{subfig:3d-mvdream}
  \end{subfigure}
  \caption{\textbf{Comparisons to baseline on text-to-3D Generaion.} Our method can generate
  diverse 3D models with realistic and detailed appearances. We compare our method with the baseline using different pretrained Diffusion Models. Prompt for this figure: ``A typewriter made out of vegetables''. See additional experiments in Appendix.
  }
  \label{fig:3d-exp}
\end{figure}
\label{sec:exp}
\subsection{Implementation Details}
Unless otherwise specified, our experiments are conducted 
with the threestudio codebase\cite{threestudio2023}. We mainly conduct text-to-3D experiments using two different categories of pretrained Diffusion Models. Specifically, Stable Diffusion\cite{song2020denoising} that could generate a single image from the text prompt and MVDream\cite{shi2023mvdream} that could generate multi-view images of an object from the text prompt. We apply several tricks that we found helpful to improve generation quality \textbf{on both baselines and our method}.
We use random seeds from 0 to 3 for each prompt by default to demonstrate the diversity of generated samples. Please refer to Appendix for more implementation details.

\subsection{Evaluation on Text-to-3D Generation}
\subsubsection{Stable Diffusion.}
We visualize our experiment results of SDS\cite{shi2023mvdream, wang2023score} (baseline) and FSD in \cref{subfig:3d-sd} using Stable Diffusion\cite{song2020denoising} (\cref{subfig:3d-sd}). Due to single-image Diffusion Models' lack of 3D knowledge, both SDS and FSD may have multi-face Janus problem\cite{poole2022dreamfusion} and can not generate plausible results for all random seeds. So the generation results are usually cherry-picked in previous works, which may be potentially unfair. Instead of cherry-picking on generated samples, we manually pick some prompts on which SDS-like methods may not suffer from multi-face Janus problem in this experiment.

\subsubsection{MVDream.}  
We visualize our experiment results of SDS \cite{shi2023mvdream, wang2023score} (baseline) and FSD (ours) using MVDream\cite{shi2023mvdream} in \cref{subfig:3d-mvdream}. From our practical observations, MVDream usually yields degraded generation results compared with Stable Diffusion\cite{song2020denoising} but seldomly suffers from multi-face Jauns problem. As a result of the degradation, text-to-3D generation results of MVDream are almost identical with different random seeds when using SDS. However, FSD can still generate diverse results with MVDream.

\subsubsection{Quantitative Results.}
Please refer to Appendix.

\begin{figure}[t]
  \centering
    \includegraphics[width=0.9\linewidth]{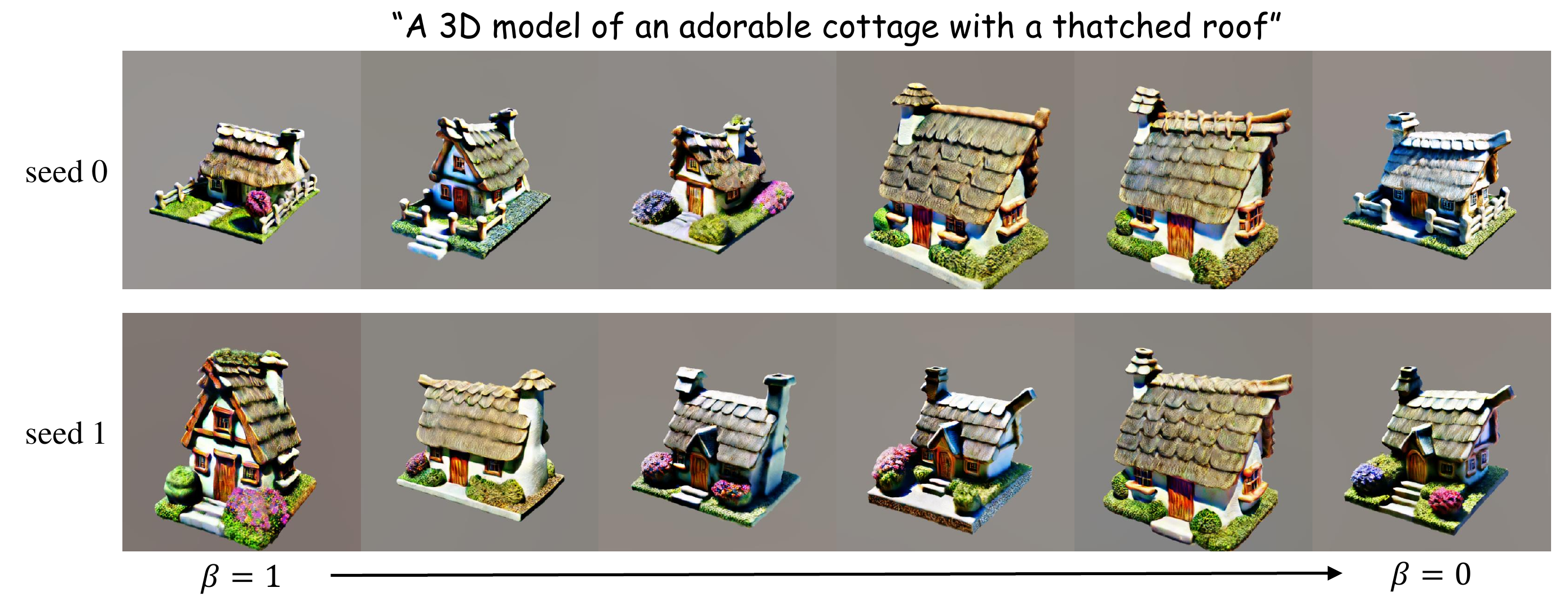}
  \caption{\textbf{Ablation on blending factor $\beta$.} parameter $\beta$ in the world-map noise function (\cref{eq:fsd-noise}) is designed for balance between FSD ($\beta=1$) and SDS ($\beta=0$). We set $\beta=1$ in the other part of this work and show results of FSD with different $\beta$s in this figure.
  }
  \label{fig:betas}
\end{figure}

\subsection{Ablation Study}
\subsubsection{Ablation on blending factor $\beta$.}
$\beta$ in \cref{eq:fsd-noise} is the blending factor designed for balancing between the deterministic noise of FSD and the random noise of SDS. In the other part of this work, we fix $\beta=1$, which means the noise function $\ec$ is completely deterministic. We show the impact of beta in \cref{fig:betas}.

\subsubsection{Ablation on other hyperparameters in $\ec$.} 
We use a parameter $\Theta$ to control the size ($H$ and $W$) of noise world-map in \cref{eq:fsd-noise}. $H$ and $W$ are determined by $\Theta$ according to $H=H_{\text{hidden}}\frac{\pi}{\Theta}$ and $W=W_{\text{hidden}}\frac{2\pi}{\Theta}$. In practice, we found FSD are prone to the parameter $\Theta$. See details in Appendix.

\section{Conclusion}
\label{sec:conslusion}
\subsubsection{Conclusions.}
In this work, we systematically study the problem of text-to-3D generation. We first review the theorems of SDS and reveal a simple but profound underlying connection between DDIM and SDS. Following this insight, we propose FSD to tackle the diversity degradation challenge. By using a consistent noise sampling schedule that aligns noise coarsely in 3D space, FSD breaks free from the maximum-likelihood-seeking nature of SDS and could generate diverse results without compromising generation quality.

\subsubsection{Limitations and future works}
Although FSD could improve the diversity of 3D generation, we found that it is still difficult to generate 3D models as diverse as the images generated through DDIM generation process. Moreover, we only found plausible theorems for FSD-guided 2D generation, and generalized FSD to 3D generation in an intuitive way. As we do not specify the noise function $\ec$ in the general form of FSD (\cref{eq:fsd}), we design $\ec$ manually to coarsely align noise prior in 3D space in this work, where better designs of $\ec$ may exist. We leave this for future work.
%
%
\bibliographystyle{splncs04}
\bibliography{main}

\clearpage
\appendix
\section*{Appendix}

\section{Quantitative Results}
We compute several metrics for 3D generation results with SDS\cite{poole2022dreamfusion, wang2023score} and FSD. The hyperparameters are kept the same as those in \cref{app-sec:additional}. When using Stable Diffusion\cite{song2020denoising} as backbone, we use 16 prompts and 4 random seeds for each prompt (\cref{app-tab:sd}). When using MVDream\cite{shi2023mvdream} as backbone, we also use 16 prompts and 4 random seeds for each prompt (\cref{app-tab:mvdream}).
\begin{table}[htb]
  \caption{Stable Diffusion\cite{song2020denoising} as backbone
  }
  \label{app-tab:sd}
  \centering
  \begin{tabular}{@{}c|ccc@{}}
    \toprule
    Method & CLIP ($\uparrow$) & IS ($\uparrow$) & CROSS-FID ($\uparrow$)\\
    \midrule
    DDIM Images 
    & $33.72\pm1.83$ & $1.68\pm0.55$ & -\\
    \midrule
    SDS\cite{poole2022dreamfusion,wang2023score} 
    & $32.57\pm1.43$ & $1.58\pm0.47$ & $106.5\pm58.3$\\
    FSD(ours) 
    & $\mathbf{32.72\pm1.56}$ & $\mathbf{1.78\pm0.49}$ & $\mathbf{141.8\pm57.9}$\\
  \bottomrule
  \end{tabular}
\end{table}

\begin{table}[htb]
  \caption{MVDream\cite{shi2023mvdream} as backbone}
  \label{app-tab:mvdream}
  \centering
  \begin{tabular}{@{}c|ccc@{}}
    \toprule
    Method & CLIP ($\uparrow$) & IS ($\uparrow$) & CROSS-FID ($\uparrow$)\\
    \midrule
    DDIM Images & 
    $34.64\pm2.56$ & $2.02\pm0.47$ & -\\
    \midrule
    SDS\cite{poole2022dreamfusion,wang2023score} & 
    $\mathbf{30.93\pm3.40}$ & $1.77\pm0.37$ & $86.6\pm33.4$\\
    FSD(ours) & 
    $30.12\pm3.07$ & $\mathbf{2.13\pm0.35}$ & $\mathbf{174.8\pm44.5}$\\
  \bottomrule
  \end{tabular}
\end{table}

\subsubsection{CLIP score}
We compute CLIP score\cite{hessel2021clipscore, radford2021learning} using ViT-B/32 to measure the semantic similarity between the renderings of the generated 3D object and the input text prompt. We sample 24 views for each prompt and each seed when computing CLIP score. 

\subsubsection{IS score}
We compute IS score\cite{salimans2016improved} to measure both the image quality and diversity. We first compute the IS scores of sampled views for each prompt and then average the IS scores across prompts.

\subsubsection{CROSS-FID score}
To directly measure the diversity of generation results, it is natural to measure the inception distance between different generated samples. We first sample 24 views for each prompt and each seed. Then we separate the images corresponding to random seeds 0, 1 and 2, 3 into two sets of images. We compute FID\cite{heusel2017gans} of the two sets of images and average the FID score across prompts. We term this score as CROSS-FID score since it is different from the standard way of using FID to evaluate GANs\cite{heusel2017gans}.

\section{Discussions}\label{app:sec:discussion}
\subsubsection{Difference with Consistent3D}
Recent work Consistent3D~\cite{wu2024consistent3d} also applied fixed noise when conducting SDS-like generation. In Consistent3D, they follow the idea of Consistent Training~\cite{song2023consistency} and use the rendered image perturbed with fixed noise to approximate the starting point of the deterministic flow. In our method, for the same camera view, we also add fixed noise to the rendered image, but the noised image is used to simulate a variable in the middle of a PF-ODE trajectory, which is different from Consistent3D. Our FSD loss is also different from the CDS loss in Consistent3D, even when our view-dependent noise function gives the same noise for all camera views, implying an essential difference between our method and Consistent3D.

\subsubsection{Connection to Signal-ODE}
Our reformulated ODE (\cref{eq:pf-ode-sds}) is equivalent to the Signal-ODE presented in the concurrent and independent work BOOT~\cite{gu2023boot}, which aims to distill a fast image generator. When the diffusion model is changed to sample prediction in \cref{eq:pf-ode-sds}, our reformulated PF-ODE is the same as the Signal-ODE in BOOT. In BOOT, they let the student image generation model predict the clean variables $\xo$ on the ODE trajectory, while our method uses images rendered from 3D representation $\theta$ to model the clean variables $\xo$ on the ODE trajectory.

\section{Additional Experiment Results}
\label{app-sec:additional}
\subsection{Additional Generation Results}
\begin{figure}[tp]
  \centering
\includegraphics[width=\linewidth]{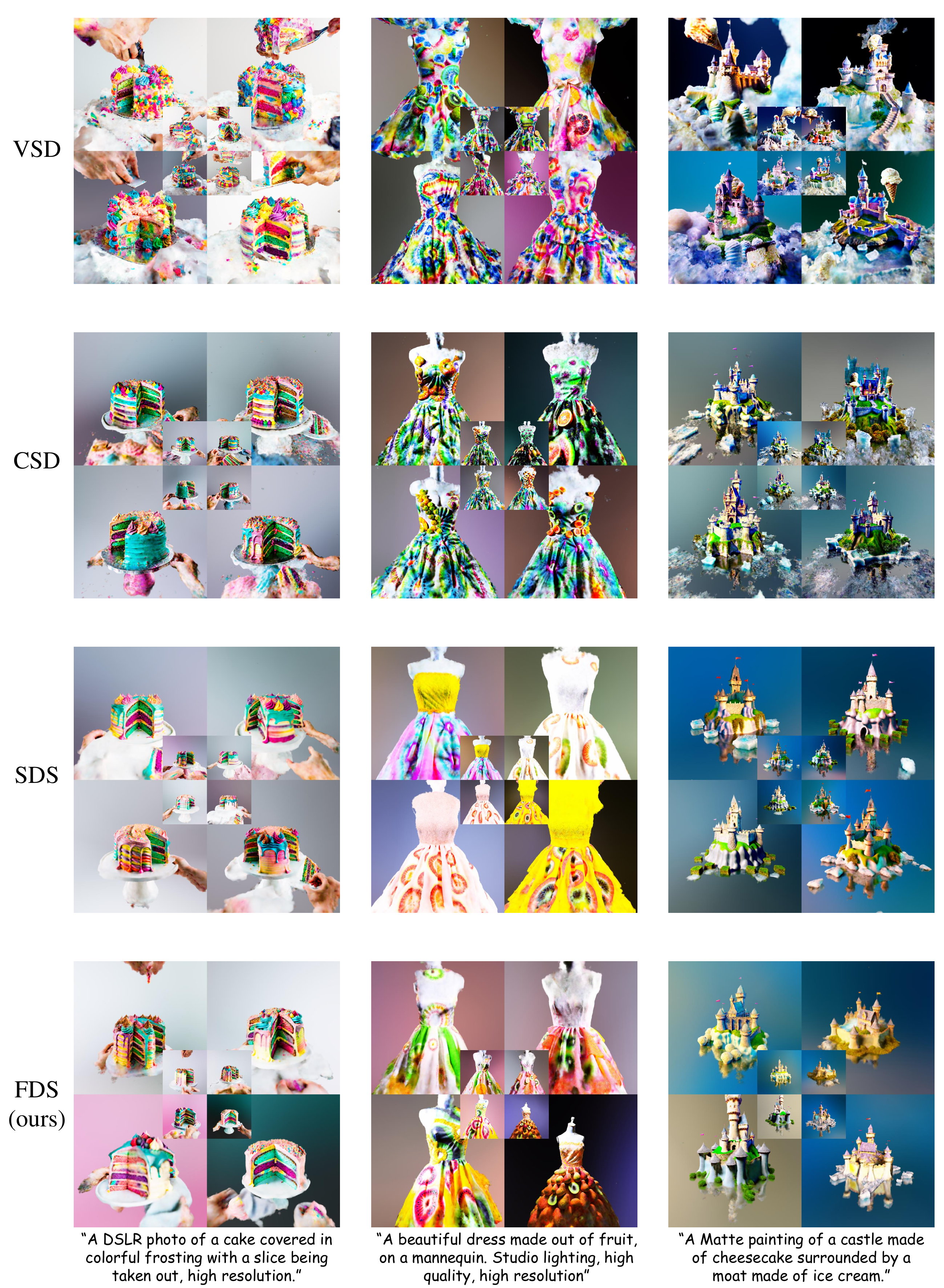}
  \caption{\textbf{Comparisons between FSD and SDS-like baseline methods.} We present results of FSD and compare them to baseline methods, namely ProlificDreamer (VSD)\cite{wang2024prolificdreamer}, SDS\cite{poole2022dreamfusion, wang2023score}, CSD\cite{ho2022classifier}. We use random seeds from 0 to 3 for each prompt.
  }
  \label{app-fig:baselines}
\end{figure}
\begin{figure}[tp]
  \centering
\includegraphics[width=\linewidth]{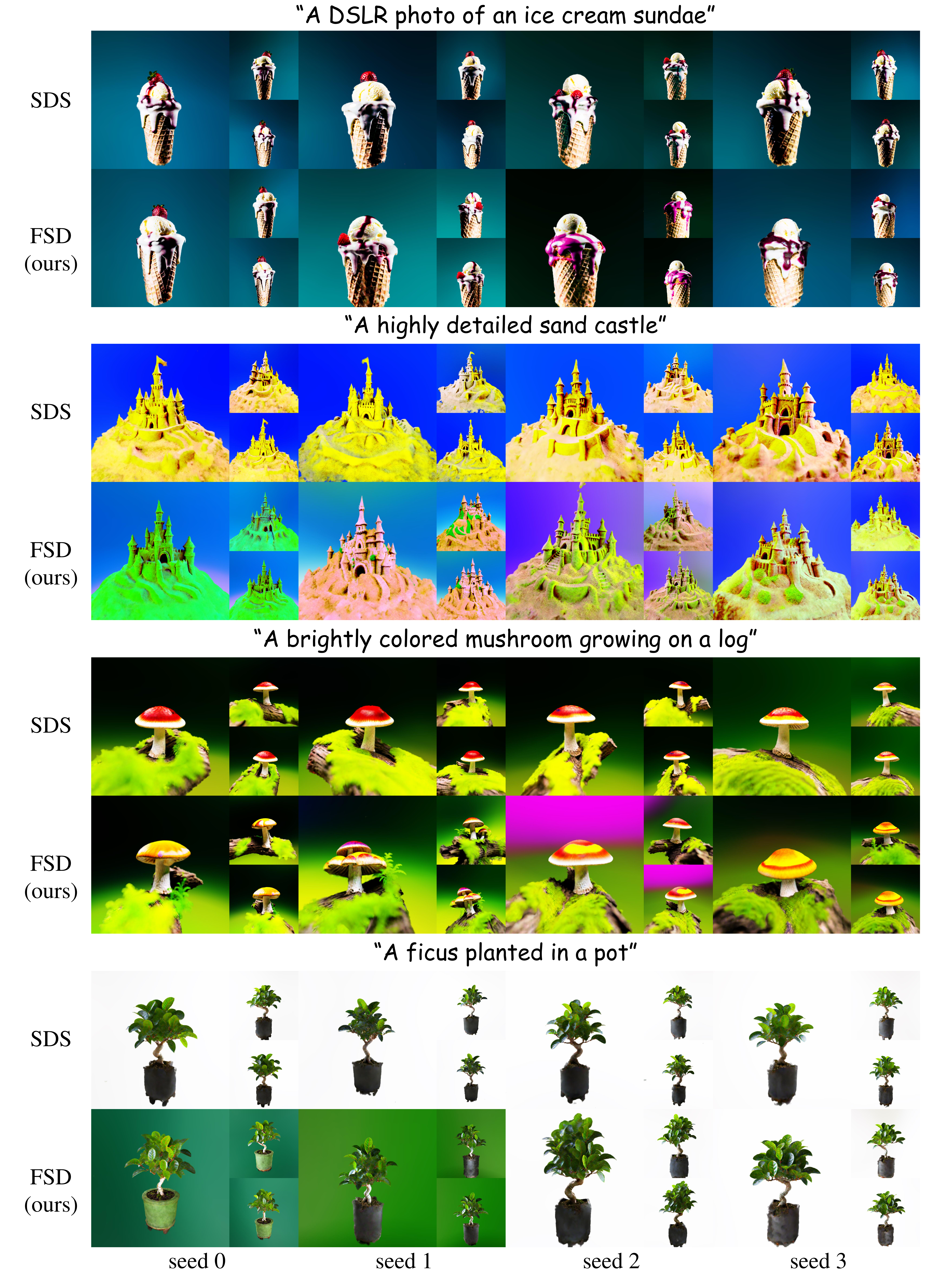}
  \caption{\textbf{Experiment results of FSD and SDS\cite{poole2022dreamfusion, song2020score} with Stable Diffusion\cite{song2020denoising} as backbone.}
  }
  \label{app-fig:sd}
\end{figure}
\begin{figure}[tp]
  \centering
\includegraphics[width=\linewidth]{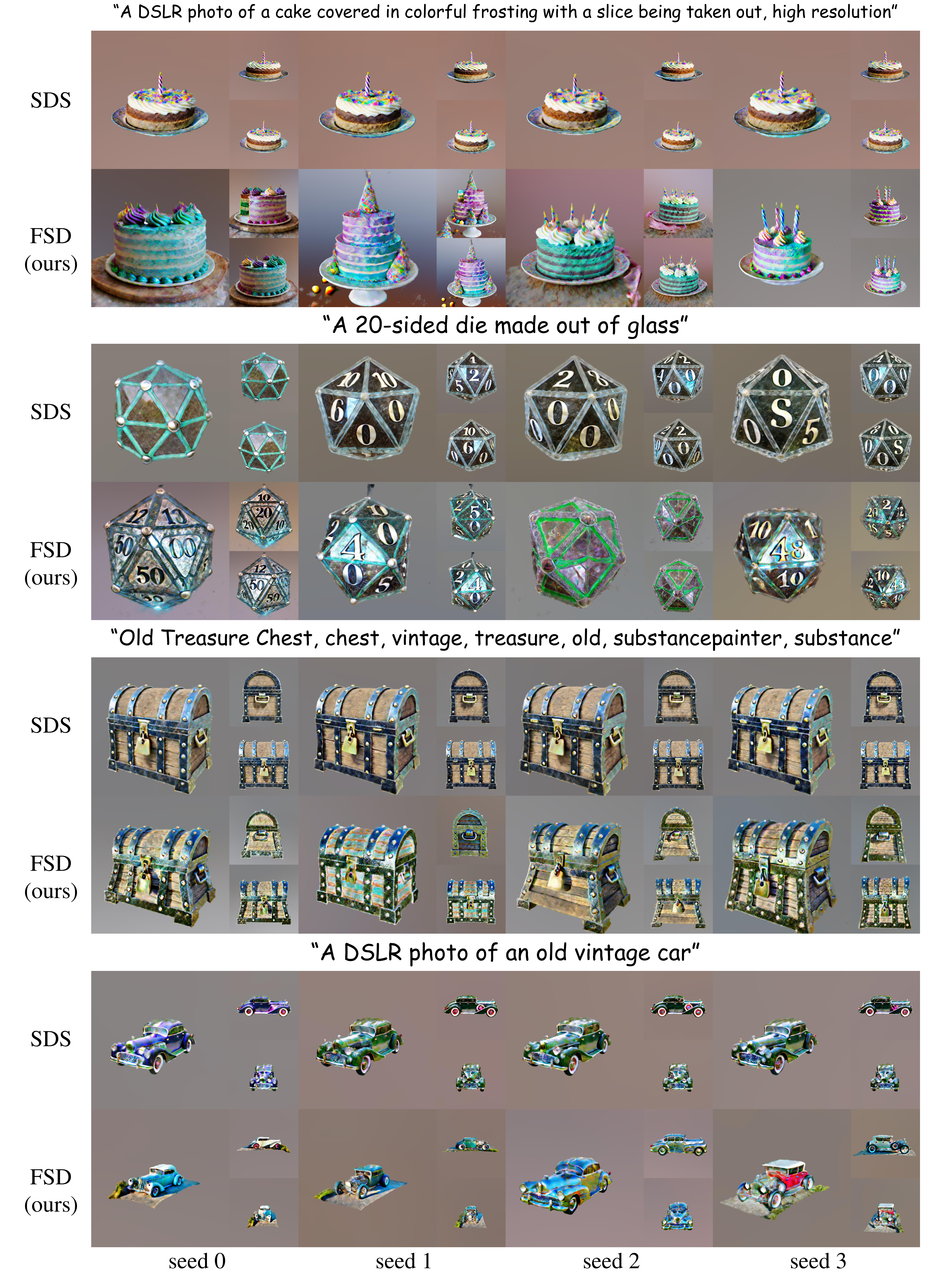}
  \caption{\textbf{Experiment results of FSD and SDS\cite{poole2022dreamfusion, song2020score} with MVDream\cite{shi2023mvdream} as backbone.}
  }
  \label{app-fig:mvdream}
\end{figure}
We provide additional 3D experiment results using FSD and other SDS-like methods in \cref{app-fig:baselines}, \cref{app-fig:sd} and \cref{app-fig:mvdream}.

Specifically, in \cref{app-fig:baselines}, we present results of FSD and compare them to baseline methods, namely ProlificDreamer (VSD)\cite{wang2024prolificdreamer}, SDS\cite{poole2022dreamfusion, wang2023score}, CSD\cite{ho2022classifier}. In \cref{app-fig:sd}, we provide additional comparisons between SDS and FSD with Stable Diffusion\cite{song2020denoising} as the backbone Diffusion Model. In \cref{app-fig:mvdream}, we provide additional comparisons between SDS and FSD with MVDream\cite{shi2023mvdream} as the backbone Diffusion Model. We do not provide experiment results with Diffusion Models designed for novel view synthesis (\eg SyncDreamer\cite{liu2023syncdreamer}) since we found FSD does not improve the diversity of the back view of an object significantly given the front view when using Diffusion Models of such category.

\subsection{Additional Ablations}

\begin{figure}[t]
  \centering
  \begin{subfigure}{0.18\linewidth}
    \includegraphics[width=\linewidth]{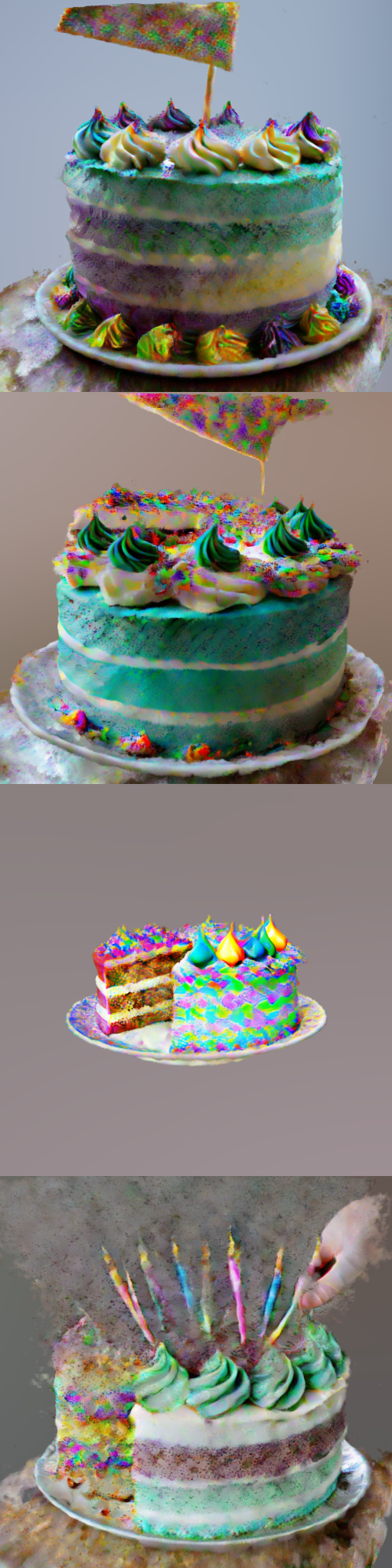}
    \caption*{$\Theta=120^{\circ}$}
  \end{subfigure}
  \hfill
  \begin{subfigure}{0.18\linewidth}
    \includegraphics[width=\linewidth]{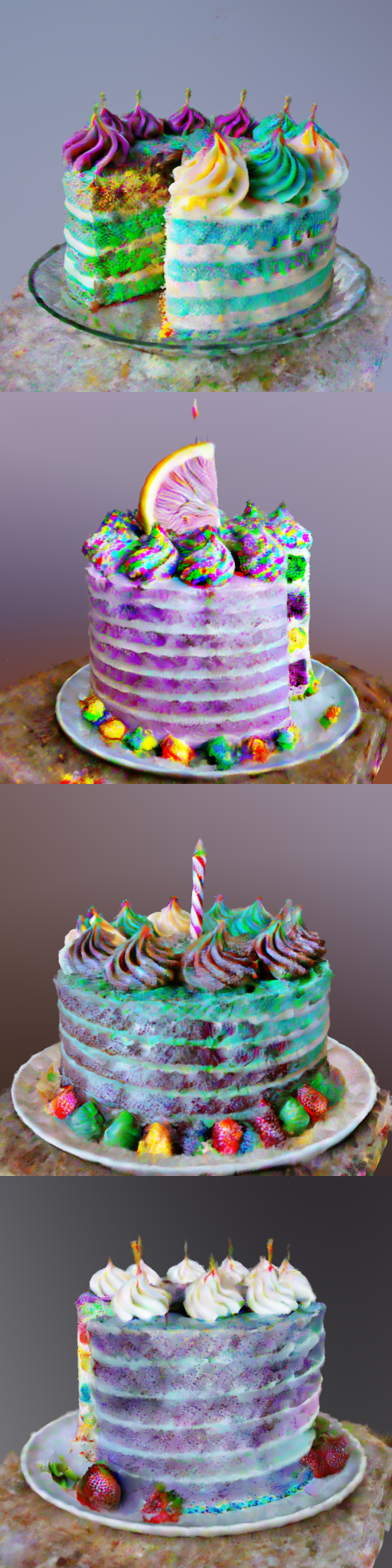}
    \caption*{$\Theta=180^{\circ}$}
  \end{subfigure}
  \hfill
  \begin{subfigure}{0.18\linewidth}
    \includegraphics[width=\linewidth]{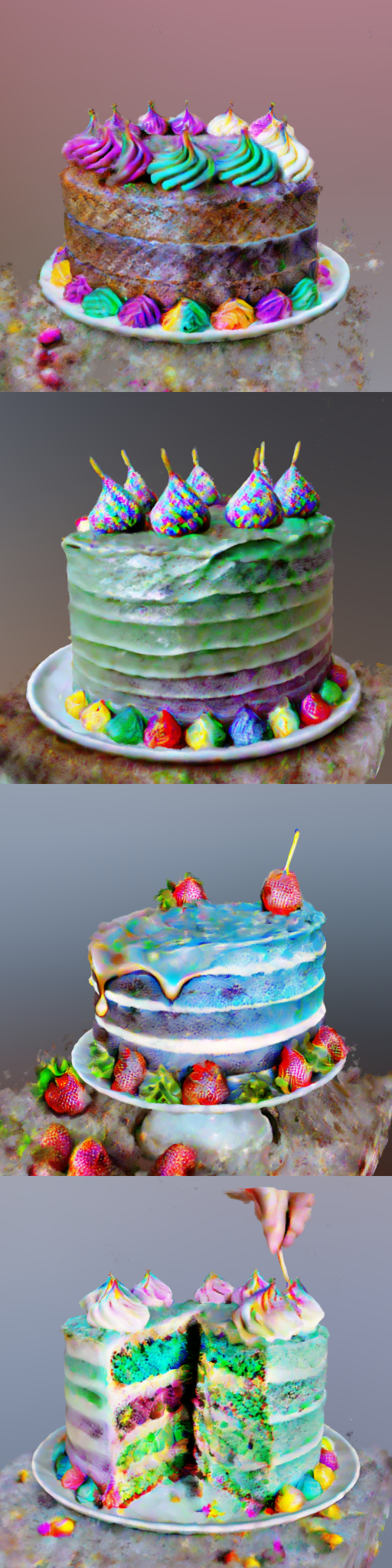}
    \caption*{$\Theta=240^{\circ}$}
  \end{subfigure}
  \hfill
  \begin{subfigure}{0.18\linewidth}
    \includegraphics[width=\linewidth]{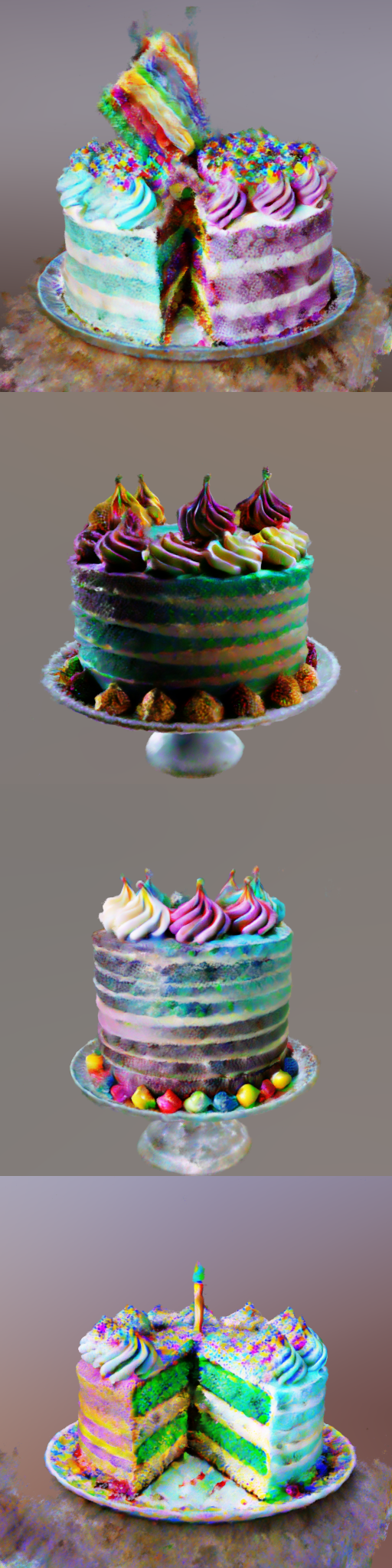}
    \caption*{$\Theta=300^{\circ}$}
  \end{subfigure}
  \hfill
  \begin{subfigure}{0.18\linewidth}
    \includegraphics[width=\linewidth]{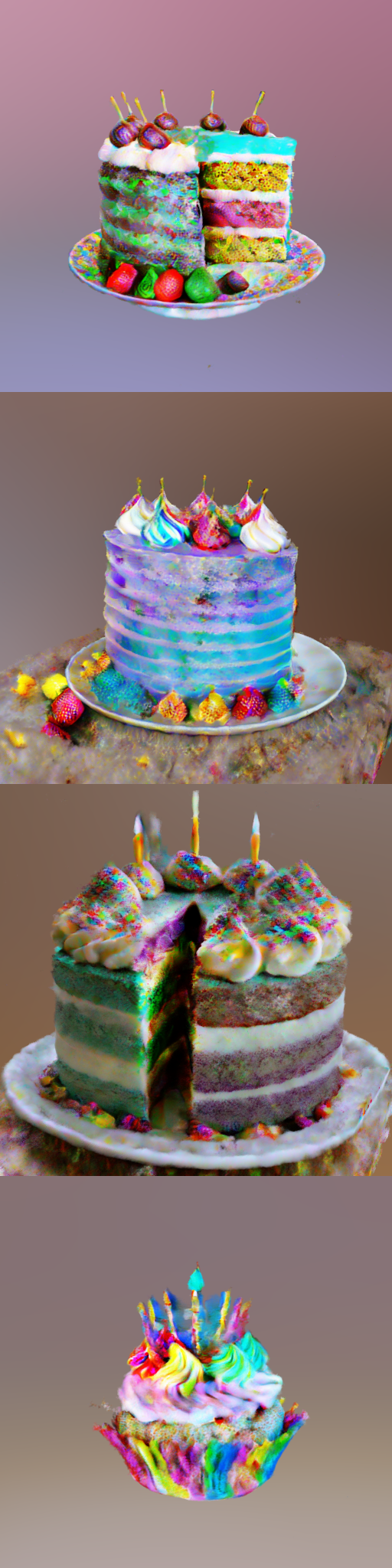}
    \caption*{$\Theta=360^{\circ}$}
  \end{subfigure}
  \caption{\textbf{Ablation on other hyperparameters in $\ec$.} We use a parameter $\Theta$ to control the size of noise world map ($H$ and $W$) in $\ec$. When $\Theta$ is larger, the ``radius $r_+$'' (\cref{app-eq:rp-theta} and \cref{app-eq:rp-phi}) of the noise world map is smaller and FSD trends to generate smaller 3D models. In practice, we found FSD is prone to the parameter $\Theta$.
  }
  \label{app-fig:thetas}
\end{figure}

We use a parameter $\Theta$ to control the noise world-map's size ($H$ and $W$). $H$ and $W$ are determined by $\Theta$ according to $H=H_{\text{hidden}}\frac{\pi}{\Theta}$ and $W=W_{\text{hidden}}\frac{2\pi}{\Theta}$. We visualize the results corresponding to different $\Theta$s in \cref{app-fig:thetas}.

\begin{figure}[t]
  \centering
  \hfill
  \begin{subfigure}{0.45\linewidth}
      \includegraphics[width=\linewidth]{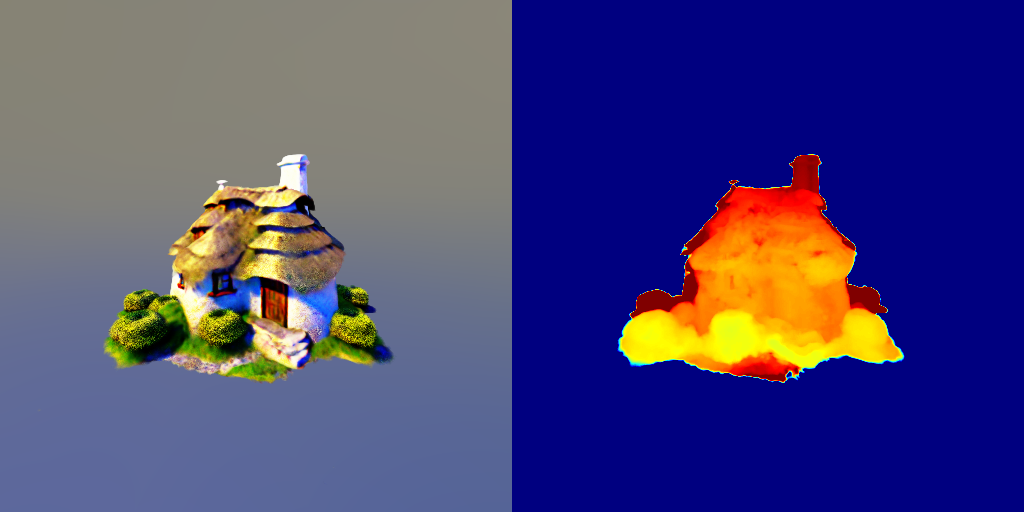}\\
      \includegraphics[width=\linewidth]{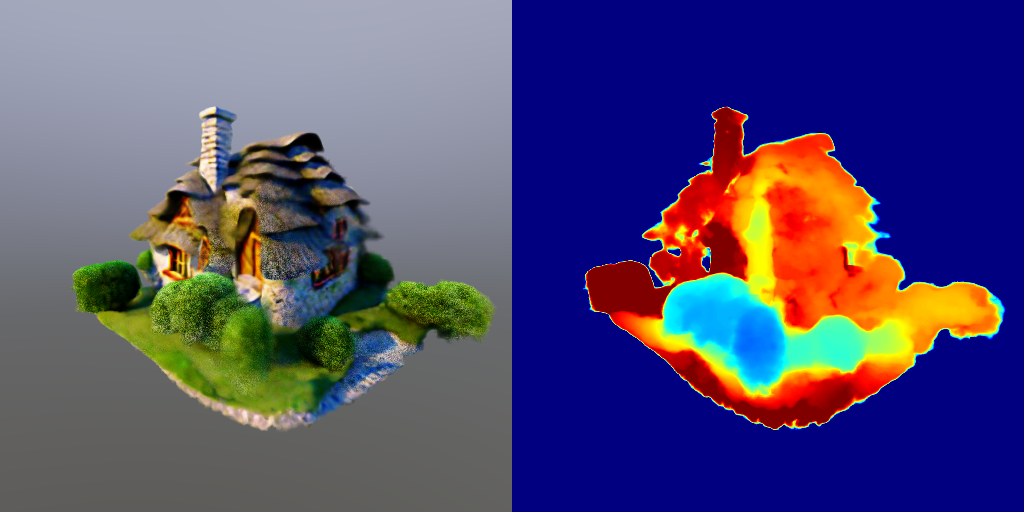}\\
      \includegraphics[width=\linewidth]{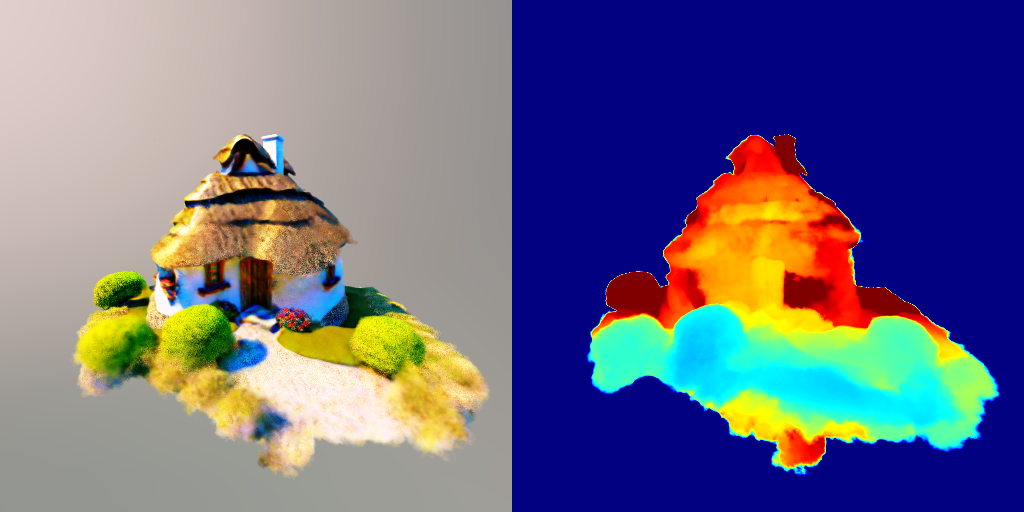}\\
      \includegraphics[width=\linewidth]{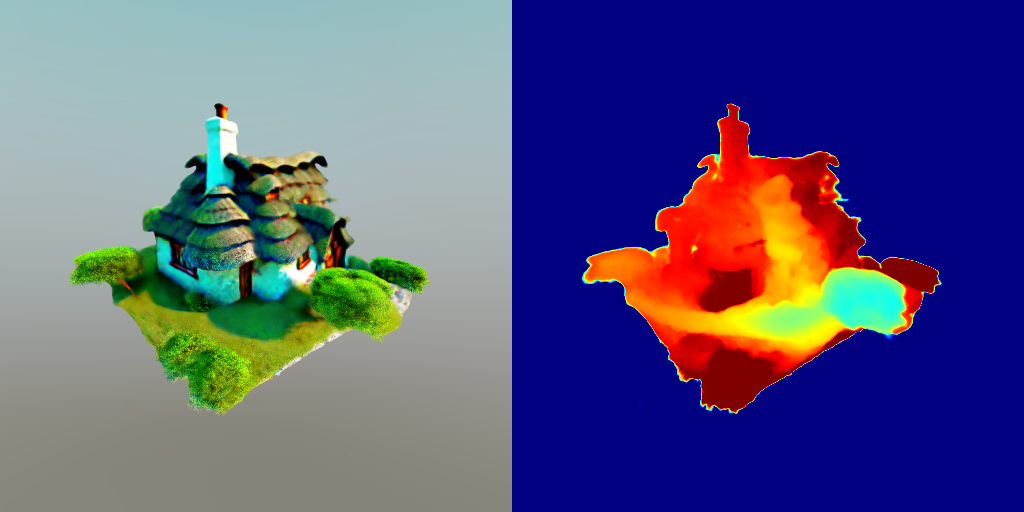}
      \caption{vanilla constant noise function}
  \end{subfigure}
  \hfill
  \begin{subfigure}{0.45\linewidth}
      \includegraphics[width=\linewidth]{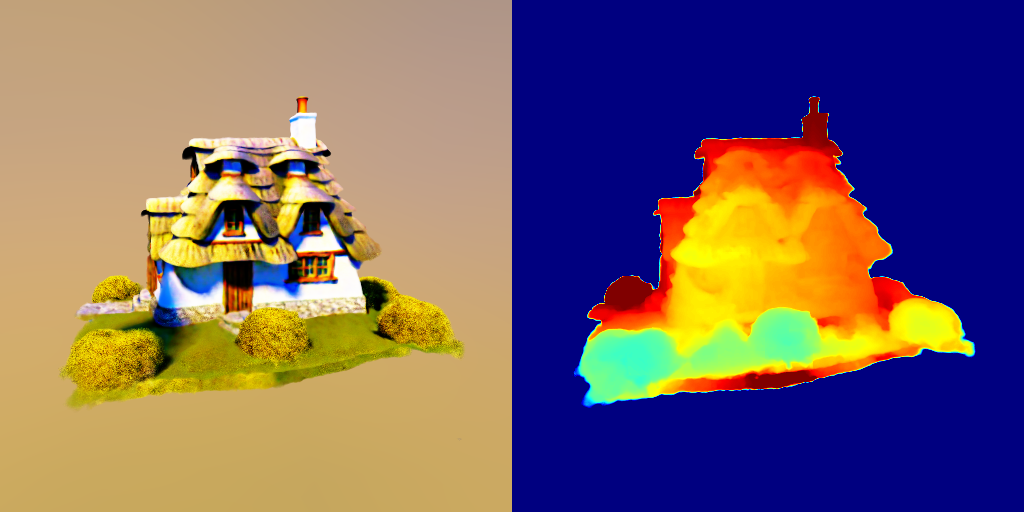}\\
      \includegraphics[width=\linewidth]{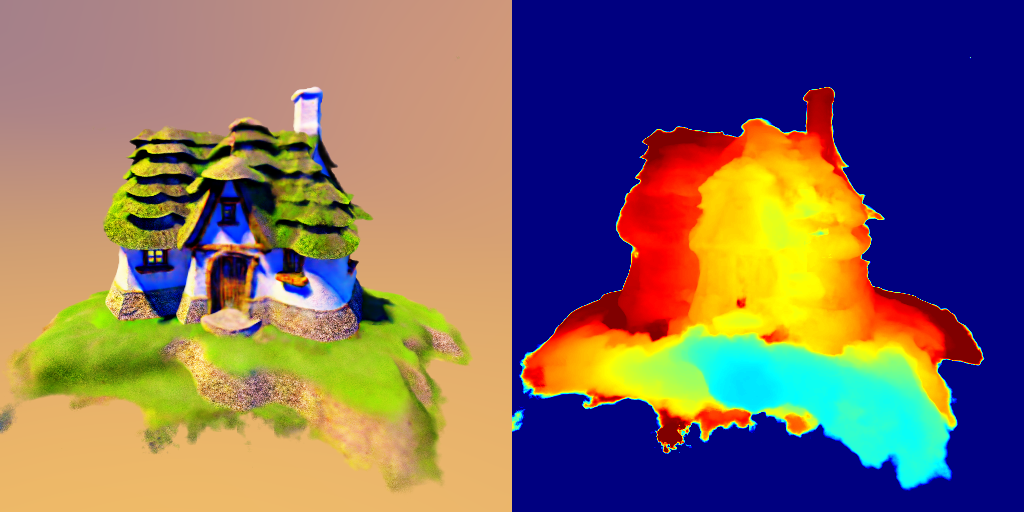}\\
      \includegraphics[width=\linewidth]{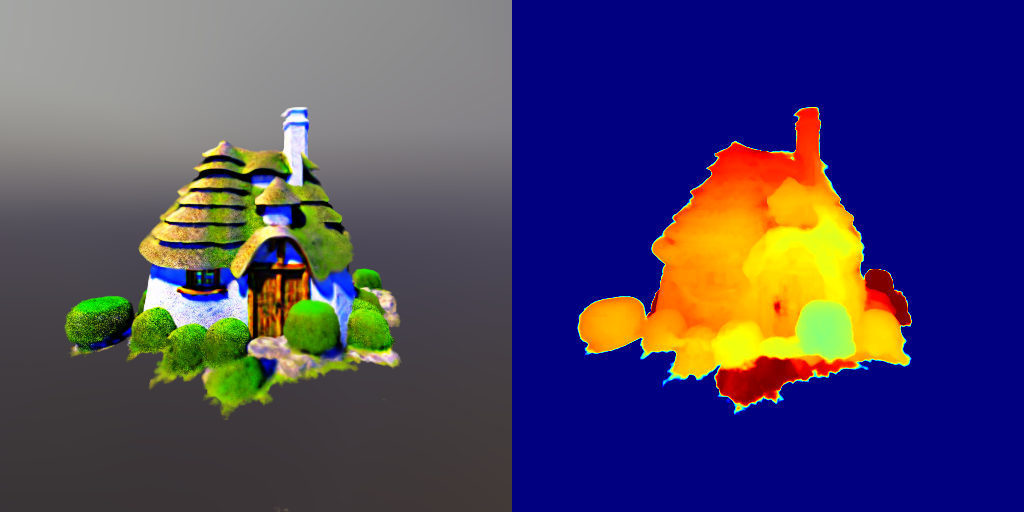}\\
      \includegraphics[width=\linewidth]{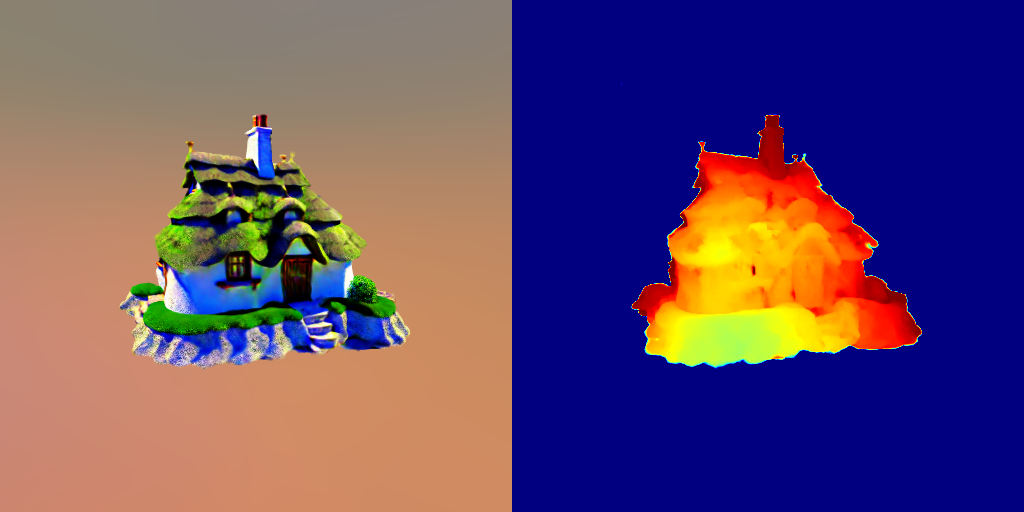}
      \caption{world-map noise function (proposed)}
  \end{subfigure}
  \hfill
  \caption{\textbf{Ablation on design of $\ec$.} We use noise function $\ec$ to align noise prior in FSD. We present RGB maps and depth maps of the generated samples using random seeds from 0 to 3. A vanilla constant function may result in holes on the surfaces. In contrast, the world-map noise function will not lead to any obvious flaws.
  }
  \label{app-fig:vanilla}
\end{figure}

We also provide an additional comparison between the vanilla constant noise function and the world-map noise function in \cref{app-fig:vanilla}. We find the constant noise function may harm the geometry of the generation results.

\subsection{Additional Experiments on Noise Prior}

\begin{figure}[tbh]
\centering
\includegraphics[width=0.95\linewidth]{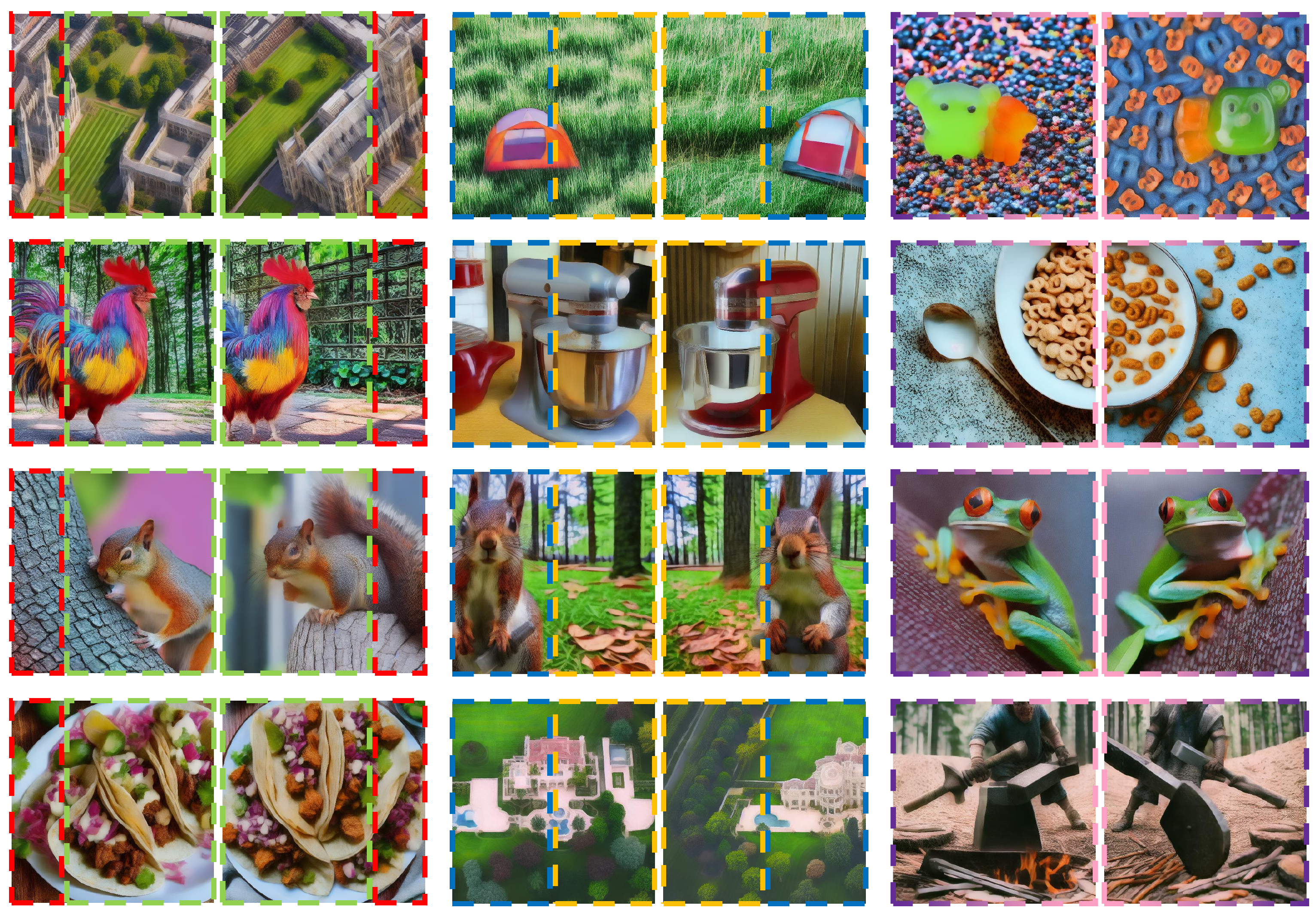}
  \caption{\textbf{Impact of initial noise $\epsf$.} We provide additional 2D experiment results generated by FSD show the impact of the initial noise $\epsf$. The patches framed with the same color in the same row share the same initial noise patches. 
  }
  \label{app-fig:fsd-2d-local}
\end{figure}
We provided additional generation results of FSD on 2D image space with shuffled or flipped initial noise in \cref{app-fig:fsd-2d-local}. The patches framed with the same color in the same row share the same initial noise patches $\epsf$.

\section{Implementation Details}
\subsection{3D Experiments using FSD}

The backbone Diffusion Model for Fig. 1 in the main text is MVDream\cite{shi2023mvdream}. The configuration for this figure is the same as Fig. 7 in the main text.

We use sqrt-annealing proposed by HiFA\cite{zhu2023hifa} for 3D experiments in this work. We use the same CFG\cite{ho2022classifier} scale for SDS\cite{shi2023mvdream, wang2023score} and FSD. We use CFG=1 when applying image loss trick\cite{zhu2023hifa} since we find the reconstructed images are unrealistic when the CFG scale is large, which is not good guidance. We follow the default setting of threestudio\cite{threestudio2023} code base for other hyperparameters. We keep the iteration number and learning rate of FSD the same as SDS\cite{shi2023mvdream, wang2023score} in 3D experiments. We mainly conduct our experiments using NeRF representation\cite{muller2022instant} since SDS-like methods are not sensitive to the form of 3D representations.

\subsection{2D Experiments using FSD}

\begin{table}[tb]
  \caption{Implementation details on 2D experiments with FSD.}
  \label{app-tab:impl-2d}
  \centering
  \begin{tabular}{@{}l|cccccc@{}}
    \toprule
    Methods Name & SDS\cite{poole2022dreamfusion,wang2023score} & NFSD\cite{katzir2023noise} & VSD\cite{wang2024prolificdreamer} & FSD(ours) & DDIM\cite{song2020denoising}\\
    \midrule
    Iteration Num & 500 & 500 & 500 & 500 & 50\\
    CFG\cite{ho2022classifier} & 100 & 7.5 & 7.5 & 7.5 & 7.5\\
    Learning Rate & 2e-2 & 2e-2 & 2e-2 & 3e-3 & -\\
    Optimizer & Adam\cite{kingma2014adam} & Adam & Adam & Adam & -\\
    Timestep Annealing & linear & linear& linear& linear& -\\
  \bottomrule
  \end{tabular}
\end{table}
Here we describe the details of the experiments on 2D images with FSD in the main text. For Fig. 3 in the main text, we show the implementation details in \cref{app-tab:impl-2d}. Below, we provide the loss functions for convenient reference.

Let denote the prediction of Diffusion Models as $\bd{\epsilon}_y^t=\bd{\epsilon}_\phi(\bd{x}_t|y,t)$ and $c$ the classifier-free-guidance (CFG)\cite{ho2022classifier} scale. Then
\begin{equation}
    \nabla_\theta L_{\text{SDS}}^\theta = \E_{\bd{\epsilon},\bd{c},t}  \left[  \left(c \cdot (\bd{\epsilon}^t_y-\bd{\epsilon}^t_\emptyset)+(\bd{\epsilon}^t_\emptyset-\bd{\epsilon})\right)\frac{\partial \gc}{\partial \theta}\right]
\end{equation}
is the loss function of SDS\cite{poole2022dreamfusion, wang2023score}, where $\emptyset$ is the empty prompt.

\begin{equation}
    \nabla_\theta L_{\text{NFSD}}^\theta = \E_{\bd{\epsilon},\bd{c},t}  \left[  \left(c \cdot (\bd{\epsilon}^t_y-\bd{\epsilon}^t_\emptyset)+(\bd{\epsilon}^t_\emptyset-\bd{\epsilon}^t_{y_{\text{neg}}})\right)\frac{\partial \gc}{\partial \theta}\right]
\end{equation}
is the loss function of NFSD\cite{katzir2023noise}, where $y_{\text{neg}}$ is a text negative prompt.

\begin{equation}
    \nabla_\theta L_{\text{VSD}}^\theta = \E_{\bd{\epsilon},\bd{c},t}  \left[  \left(c \cdot (\bd{\epsilon}^t_y-\bd{\epsilon}^t_\emptyset)+(\bd{\epsilon}^t_\emptyset-\bd{\epsilon}^t_{\text{lora}})\right)\frac{\partial \gc}{\partial \theta}\right]
\end{equation}
is the loss function of VSD\cite{wang2024prolificdreamer}, where $\bd{\epsilon}^t_{\text{lora}}$ the Diffusion Model fine-tuned by LoRA\cite{hu2021lora}. 

\begin{equation}
    \nabla_\theta L_{\text{FSD}}^\theta = \E_{\bd{c},t}  \left[  \left(c \cdot (\bd{\epsilon}^t_y-\bd{\epsilon}^t_\emptyset)+(\bd{\epsilon}^t_\emptyset-\bd{\epsilon}(\bd{c}))\right)\frac{\partial \gc}{\partial \theta}\right]
\end{equation}
is the loss function of FSD. Additionally, $\bd{x}_t = \alpha_t \gc + \sigma_t \ec$ for FSD.

Our re-implementation of $L_{NFSD}$ is slightly different from the original\cite{katzir2023noise} design of NFSD by ignoring the condition when $t<200$, to keep the formulation simple. In this way, we find $\bd{\epsilon}^t_{\text{lora}}$ used in VSD\cite{wang2024prolificdreamer} may work in a similar way as $\bd{\epsilon}^t_{y_{\text{neg}}}$ in NFSD\cite{katzir2023noise}. As a result, single-particle VSD may not be able to generate diverse samples since it is still mode-seeking, aligning with the observation of ESD\cite{wang2023taming}. But we do not conduct further investigations which are out of the scope of our work.

\section{Theory of Flow Score Distillation}
We will provide some additional preliminaries and proof of Proposition 1 in the main text in this section.
\subsection{Diffusion PF-ODE}
The Diffusion PF-ODE \cite{song2020score} can be written in the following form:
\begin{equation}
  \frac{\dd \bd{x}_t}{\dd t} = f(t) \bd{x}_t - \frac{1}{2}g^2(t) \nabla_{\bd{x}} \log p_t(\bd{x}_t|y),\quad
  \bd{x}_T \sim p_T(\bd{x}_T),\label{eq:pf-ode}
\end{equation}
where $f(t) = \frac{\dd \log \alpha_t}{\dd t},\;
g^2(t) = \frac{\dd \sigma_t^2}{\dd t} - 2 \frac{\dd \log \alpha_t}{\dd t} \sigma^2_t
$, according to DPM-solver\cite{lu2022dpm}. To get Eq. (5) in the main text, we take the derivative of $\bd{x}_t/\alpha_t$:
\begin{equation}
\begin{aligned}
  \frac{\dd (\bd{x}_t/\alpha_t)}{\dd t}
  =& \frac{1}{\alpha_t}\frac{\dd \bd{x}_t}{\dd t} - \frac{\bd{x}_t}{\alpha_t^2}\frac{\dd \alpha_t}{\dd t}\\
  =& \frac{1}{\alpha_t}\frac{\dd \bd{x}_t}{\dd t} - \frac{\bd{x}_t}{\alpha_t}f(t)\\
  =& -\frac{g^2(t)}{2\alpha_t} \nabla_{\bd{x}} \log p_t(\bd{x}_t|y)\\
  =& -\frac{\frac{\dd \sigma_t^2}{\dd t} - 2 \frac{\dd \log \alpha_t}{\dd t} \sigma^2_t}{2\alpha_t} \nabla_{\bd{x}} \log p_t(\bd{x}_t|y)\\
  =& - (\frac{1}{\alpha_t}\frac{\dd \sigma_t}{\dd t}-\frac{\sigma_t}{\alpha_t^2}\frac{\dd \alpha_t}{\dd t})\sigma_t\nabla_{\bd{x}} \log p_t(\bd{x}_t|y)\\
  =& \frac{\dd (\sigma_t/\alpha_t)}{\dd t} \left( -\sigma_t \nabla_{\bd{x}} \log p_t(\bd{x}_t|y) \right)\\
  =& \frac{\dd (\sigma_t/\alpha_t)}{\dd t} \bd{\epsilon}_\phi(\bd{x}_t|t, y).
  \label{app-eq:pf-ode-simple}
\end{aligned}
\end{equation}
This equation is the scaled version of Diffusion PF-ODE under the notation of Karras \etal\cite{karras2022elucidating}.
\subsection{DDIM Sampling}
We derive the DDIM\cite{song2020denoising} sampling algorithm in this section. According to the first order discretization of \cref{app-eq:pf-ode-simple}:
\begin{equation}
 \frac{\bd{x}_t}{\alpha_t} - \frac{\bd{x}_s}{\alpha_s} 
 = (\frac{\sigma_t}{\alpha_t} - \frac{\sigma_s}{\alpha_s})
 \bd{\epsilon}_\phi(\bd{x}_s|s, y), \label{app-eq:ddim-scale}
\end{equation}
the sampling algorithm of DDIM\cite{song2020denoising} can be derived as the following equation:
\begin{equation}
  \bd{x}_t = \alpha_t(\frac{\bd{x}_s-\sigma_s \bd{\epsilon}_\phi (x_s|y,s)}{\alpha_s}) + \sigma_t \bd{\epsilon}_\phi (\bd{x}_s|y,s). \label{app-eq:ddim}
\end{equation}

\subsection{Proof of Proposition 1}
We present a detailed derivation of Proposition 1 in the main text in this section. We define
\begin{equation}
  \bd{x}_t = \alpha_t \xo + \sigma_t \epsf \label{app-eq:change-of-variable},
\end{equation}
for the reverse diffusion process and then we can apply change-of-variable on $\bd{x}_t$. Finally
\begin{align*}
  \frac{\dd \xo}{\dd t}
  =& \frac{\dd \frac{\bd{x}_t - \sigma_t \epsf}{\alpha_t}}{\dd t}\\
  =& \frac{\dd (\bd{x}_t/\alpha_t)}{\dd t} - \frac{\dd (\sigma_t/\alpha_t)}{\dd t} \epsf\\
  =& \frac{\dd (\sigma_t/\alpha_t)}{\dd t} (\bd{\epsilon}_\phi(\bd{x}_t|t, y)-\epsf).
\end{align*}
We also derive first-order discretization of FSD for 2D generation from \cref{app-eq:ddim-scale}:
\begin{equation}
 \xo - \hat{\bd{x}}_s^o = (\frac{\sigma_t}{\alpha_t} - \frac{\sigma_s}{\alpha_s})
 (\bd{\epsilon}_\phi(\bd{x}_s|s, y)-\epsf), \label{app-eq:fsd-1st}
\end{equation}

\section{Visualize the Change-of-Variable}

\begin{figure}[pt]
  \centering
  \hfill
  \begin{subfigure}{0.45\linewidth}
    \includegraphics[width=\linewidth]{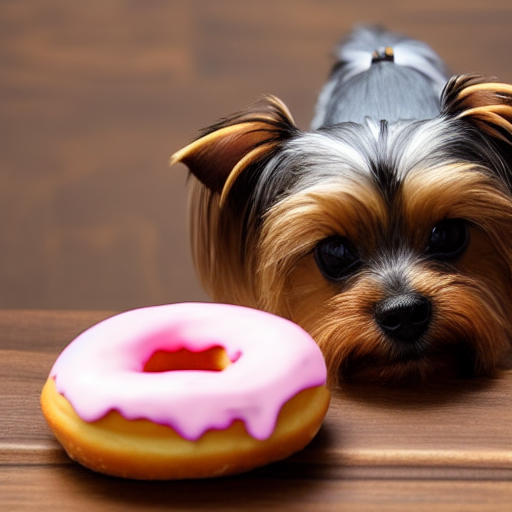}\\
    \includegraphics[width=\linewidth]{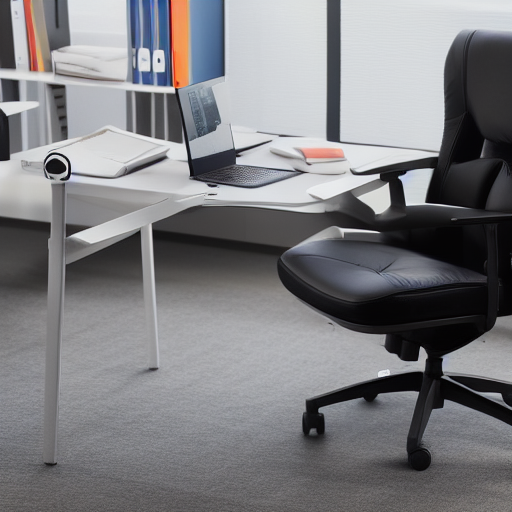}\\
    \includegraphics[width=\linewidth]{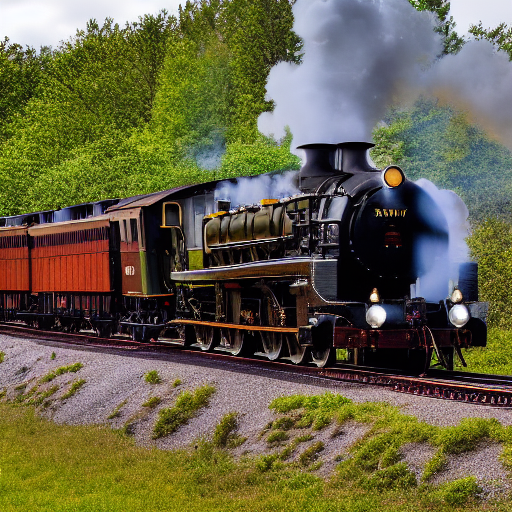}
    \caption{FSD (first-order discretization)}
  \end{subfigure}
  \hfill
  \begin{subfigure}{0.45\linewidth}
    \includegraphics[width=\linewidth]{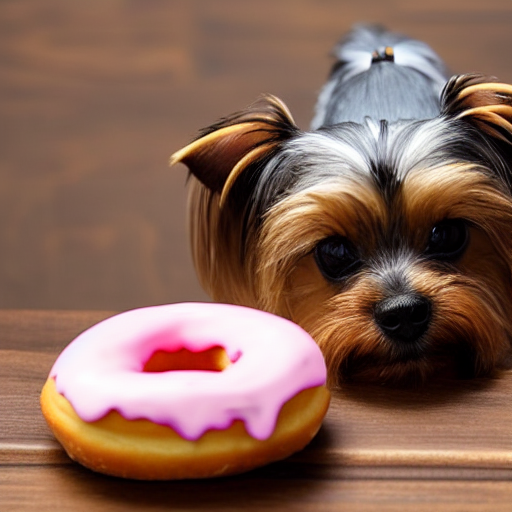}\\
    \includegraphics[width=\linewidth]{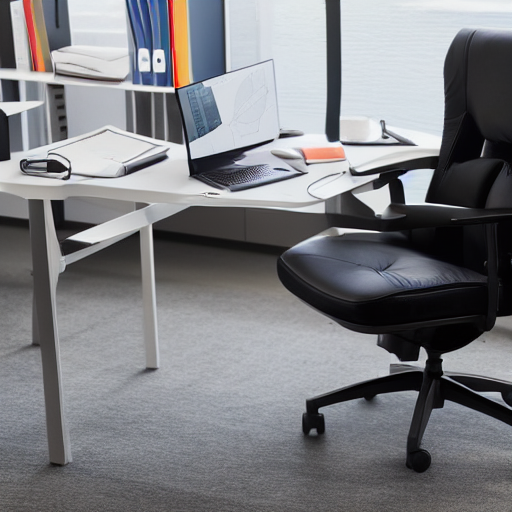}\\
    \includegraphics[width=\linewidth]{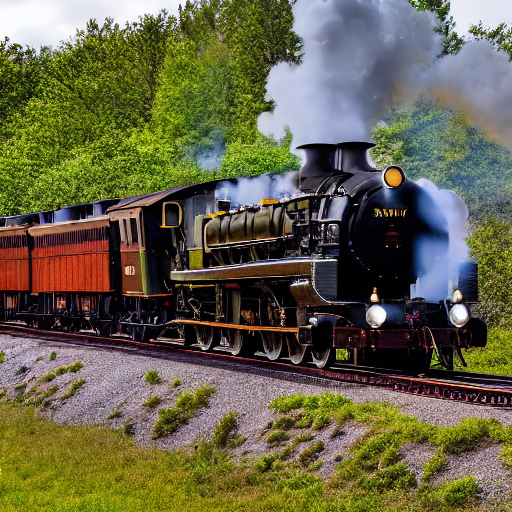}
    \caption{DDIM\cite{song2020denoising}}
  \end{subfigure}
  \hfill
  \caption{\textbf{Generation results of FSD and DDIM.} We apply FSD on 2D image generation using first-order discretization \cref{app-eq:fsd-1st} instead of Adam\cite{kingma2014adam}. In this case, we find FSD is the same as DDIM\cite{song2020denoising} except some negligible differences, which may come from the differences on handling initial conditions.
  }
  \label{app-fig:fsd-2d}
\end{figure}
Even though we show FSD can generate similar images When using the same initial noise in the main text, there are notable differences between the generated images since FSD uses Adam\cite{kingma2014adam} to update parameters while DDIM uses first-order discretization of Diffusion PF-ODE. We show generation results of FSD that use first-order discretization in \cref{app-fig:fsd-2d}.

We also visualize the ``\textit{change-of-variable}'' trick we used in the derivation of the main theorem. We define our new variable: the \textit{clean image} $\xo$ according to the following equation:

\begin{equation}
  \bd{x}_t = \alpha_t \xo + \sigma_t \epsf \label{add-eq:change-of-variable}.
\end{equation}

Notably, there is also another similar but different concept: the \textit{one-step estimated ground-truth image} $\xc$, defined by:

\begin{equation}
  \bd{x}_t = \alpha_t \xc + \sigma_t \bd{\epsilon}_\phi(\bd{x}_t|y,t) \label{add-eq:predicted-clean}.
\end{equation}

\begin{figure}[tb]
  \centering
  \begin{subfigure}{0.95\linewidth}
    \includegraphics[width=\linewidth]{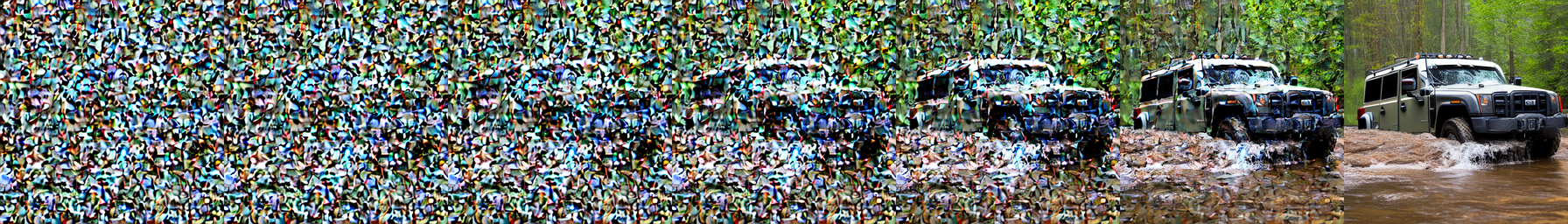}
    \caption{\textit{noisy image} $\bd{x}_t$}
  \end{subfigure}\\
  \begin{subfigure}{0.95\linewidth}
    \includegraphics[width=\linewidth]{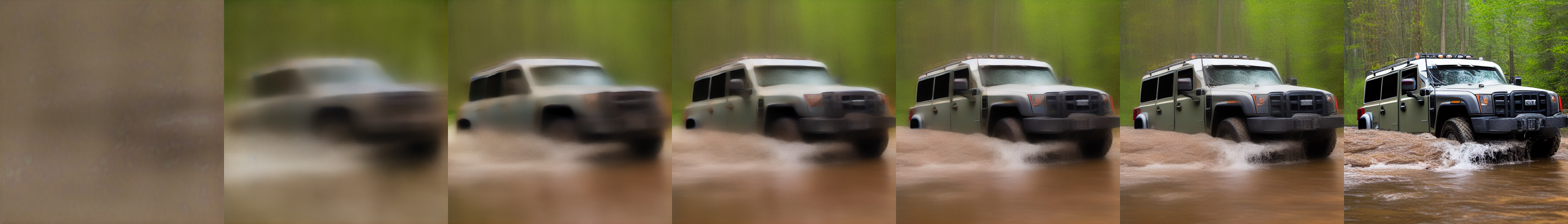}
    \caption{\textit{clean image} $\xo$}
  \end{subfigure}\\
  \begin{subfigure}{0.95\linewidth}
    \includegraphics[width=\linewidth]{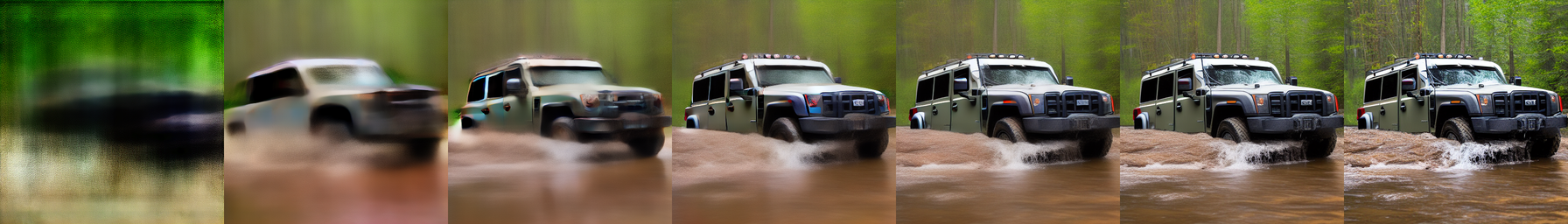}
    \caption{\textit{one-step estimated ground-truth image} $\xc$}
  \end{subfigure}\\
  \caption{\textbf{Visualization of $\bd{x}_t$, $\xo$ and $\xc$ in DDIM generation process.} $\xo$ is different from $\xc$. Moreover, one can see DDIM generation process as aligning $\xo$ with $\xc$ since $\xc-\xo \propto \nabla_\theta\Lfsd$ 
  }
  \label{app-fig:vars}
\end{figure}
\begin{figure}[htp]
  \centering
  \begin{subfigure}{0.95\linewidth}
    \includegraphics[width=\linewidth]{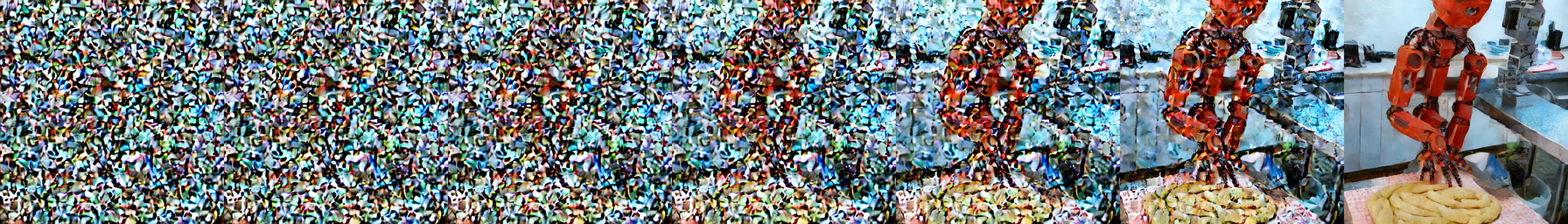}
    \caption{\textit{noisy image} $\bd{x}_t$}
  \end{subfigure}\\
  \begin{subfigure}{0.95\linewidth}
    \includegraphics[width=\linewidth]{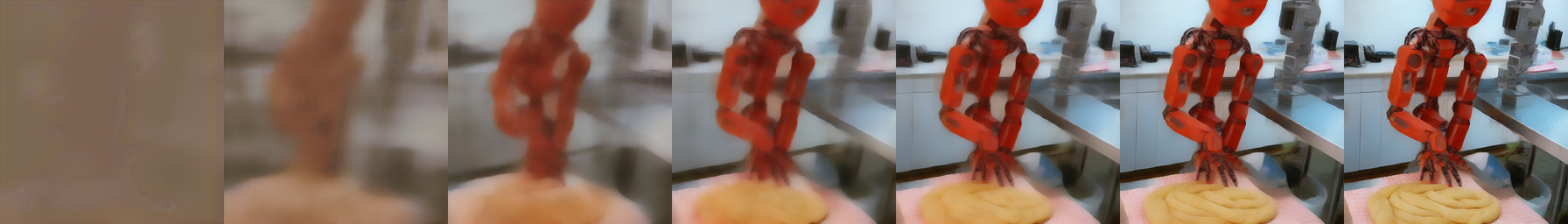}
    \caption{\textit{clean image} $\xo$}
  \end{subfigure}\\
  \begin{subfigure}{0.95\linewidth}
    \includegraphics[width=\linewidth]{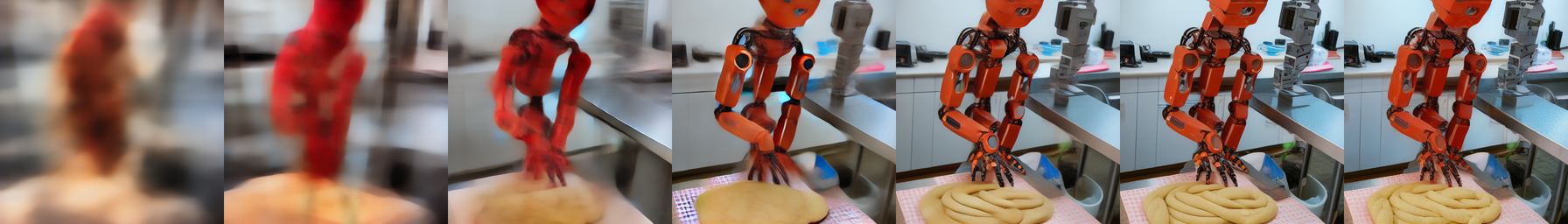}
    \caption{\textit{one-step estimated ground-truth image} $\xc$}
  \end{subfigure}\\
  \caption{\textbf{Visualization of $\bd{x}_t$, $\xo$ and $\xc$ in FSD generation process.} Compared to DDIM, FSD uses Adam\cite{kingma2014adam} to update parameters. As a result, FSD needs more iterations to converge (See \cref{app-tab:impl-2d}).
  }
  \label{app-fig:vars-fsd}
\end{figure}
\begin{figure}[htp]
  \centering
  \begin{subfigure}{0.95\linewidth}
    \includegraphics[width=\linewidth]{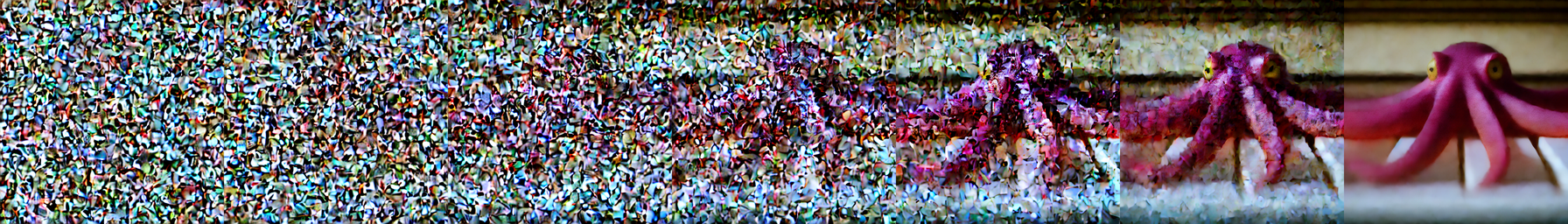}
    \caption{\textit{noisy image} $\bd{x}_t$}
  \end{subfigure}\\
  \begin{subfigure}{0.95\linewidth}
    \includegraphics[width=\linewidth]{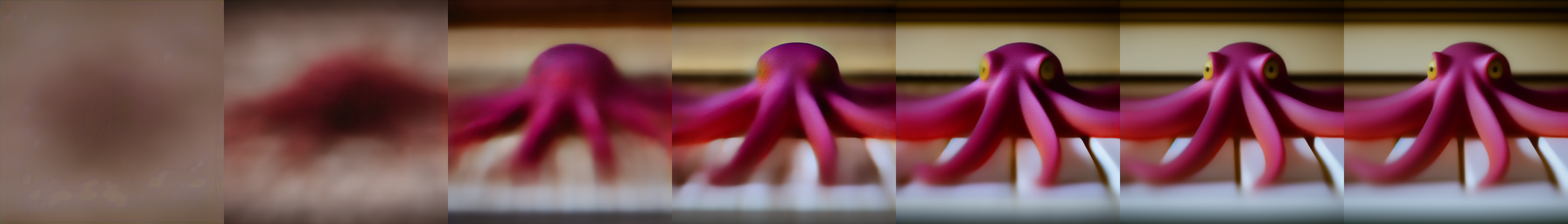}
    \caption{\textit{clean image} $\xo$}
  \end{subfigure}\\
  \begin{subfigure}{0.95\linewidth}
    \includegraphics[width=\linewidth]{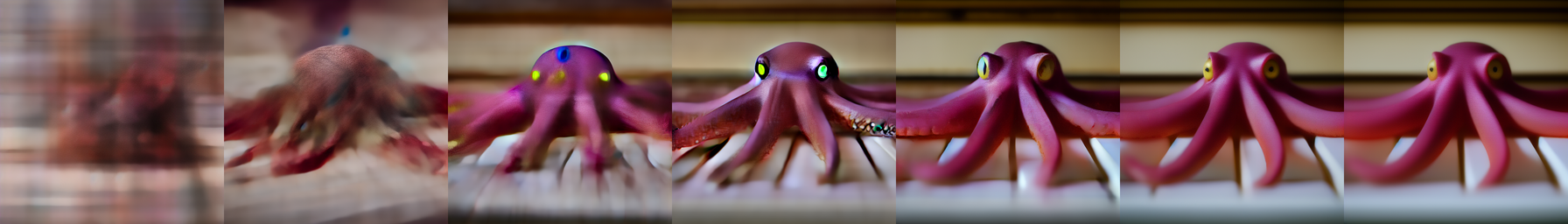}
    \caption{\textit{one-step estimated ground-truth image} $\xc$}
  \end{subfigure}\\
  \caption{\textbf{Visualization of $\bd{x}_t$, $\xo$ and $\xc$ in SDS generation process.} Compared to FSD, SDS adds random noise instead of fixed noise. As a result, the estimated GT images vary a lot and SDS needs a larger learning rate to converge (See \cref{app-tab:impl-2d}). 
  }
  \label{app-fig:vars-sds}
\end{figure}
We visualize the trajectory of $\bd{x}_t$, $\xo$ and $\xc$ in the same DDIM generation process in \cref{app-fig:vars}. As $\xc-\xo = \frac{\sigma_t}{\alpha_t} (\epsf-\bd{\epsilon}_\phi(\bd{x}_t|y,t)) \propto \nabla_\theta\Lfsd$ implies, we can see DDIM as a process that tries to align \textit{clean image} with the \textit{one-step estimated ground-truth image} generated by the Diffusion Model. We also visualize $\bd{x}_t$, $\xo$ and $\xc$ of FSD and SDS in \cref{app-fig:vars-fsd} and \cref{app-fig:vars-sds}, respecitively.

\section{Algorithm for Flow Score Distillation}
We provide a summarized algorithm for Flow Score Distillation in \cref{app-alg:fsd}.

\begin{algorithm}[H]
\begin{algorithmic}[1]
\STATE \textbf{Input:} Text-to-image Diffusion Model $\bd{\epsilon}_\phi$ and prompt $y$. Learning rate $\eta$ for parameters of the 3D representation. A monotonically decreasing function $t(\tau)$. Blending factor $\beta$.
\STATE Compute $\bd{\epsilon}_b \sim \Normal$ and $\bd{\epsilon}_p \sim \Normal$.
\FOR{$\tau \in [0, \tau_{\text{end}}]$}
    \STATE Randomly sample camera parameter $\bd{c}$.
    \STATE Render image $\gc$ and opacity mask $\bd{M}$ from 3D representation $\theta$.
    \STATE Randomly sample $\bd{\epsilon}\sim\Normal$.
    \STATE $\ec \leftarrow \sqrt{\beta} \cdot \left( (1-\bd{M}) \odot \bd{\epsilon}_b + \bd{M} \odot \window(\bd{\epsilon}_p) \right) + \sqrt{1-\beta} \cdot \bd{\epsilon}$.
    \STATE $\theta \leftarrow \theta - \eta \cdot (\bd{\epsilon}_\phi(\bd{x}_t|y,t) - \ec)\frac{\partial \gc}{\partial \theta}$
\ENDFOR
\end{algorithmic}

\caption{Flow Score Distillation}
\label{app-alg:fsd}
\end{algorithm}

\section{Practical Designing Rules for $\bd{\epsilon}(\bd{c})$}
\subsection{Practical Designing Rules}
We do not specify the form of $\ec$ in the general form of FSD in the main text. However, it is intuitive to align $\ec$ in 3D space like the noise priors used in Video Diffusion Models\cite{chang2024how, ge2023preserve, qiu2023freenoise}. We have tried several designs of $\ec$ and summarized several design rules for designing $\ec$ as well as the related potential problems if violating the design rules for $\ec$:

\begin{itemize}
    \item The noise associated with each camera view conditioning on camera view should be an uncorrelated Gaussian noise. \ie $\ec|\bd{c}\sim\Normal$. If not so, the input image may be out-of-distribution for the pretrained text-to-image Diffusion Models.
    \item The noise associated with different camera views should be aligned in 3D space. Otherwise, FSD may degrade to the original SDS. This is consistent with the observation of recent work\cite{ge2023preserve} on Video Generation, which showed that the noise maps corresponding to different video frames are highly correlated.
    \item The noise should not be only aligned in specific points in the 3D space. This may lead to broken geometry since the convergence speed in 3D space is not uniform (see analysis on failure of constant noise function in the main text). 
\end{itemize}

\subsection{World-map Noise Function}
We will take a deeper analysis of the property of the world map noise function in the main text. We will discuss how our proposed design rules are followed.

As noted in the main text, the local texture of the added noise is highly correlated with the generated images. In 3D spaces, it is natural to identify points, that correspond to similar points on the noise maps when projected onto the image planes of different camera views $\bd{c}$, as the points of high convergence speed. Let us denote those points as $\bd{p}_+$. For simplicity, we study the $\bd{p}_+$ that corresponds to the center points in image space.

For object-centric generation, a common camera view sampling strategy is to sample camera positions at $(r_{\text{cam}},\theta_{\text{cam}}, \phi_{\text{cam}})$ under spherical coordinate and force the camera to look at the center point. When we apply a constant noise function as $\ec$, the center point in 3D space is a $\bd{p}_+$ point. As a result, FSD trends to generate geometry with holes. When applying world-map noise function proposed in main text, $\bd{p}_+$ is located at $(r_+,\theta_{\text{cam}}, \phi_{\text{cam}})$. One can compute that 
\begin{equation}
    r^{\theta_{\text{cam}}}_+\approx\frac{2 \tan \frac{\text{FOV}}{2}}{2 \tan \frac{\text{FOV}}{2} + \Theta}\cdot r_{\text{cam}} \label{app-eq:rp-theta}
\end{equation}
if consider nearby views with slightly different $\theta_{\text{cam}}$ and 
\begin{equation}
    r^{\phi_{\text{cam}}}_+\approx\frac{2 \tan \frac{\text{FOV}}{2}}{2 \tan \frac{\text{FOV}}{2} + \Theta \cdot \sin \theta_{\text{cam}}}\cdot r_{\text{cam}}  \label{app-eq:rp-phi}
\end{equation}
if consider views with slightly different $\phi_{\text{cam}}$. In this way, 
we align nearby views coarsely. Moreover, for different camera parameters, $r_+$ is different, avoiding nonuniform convergence speed in 3D space.

\end{document}